\pdfoutput=1
\documentclass[11pt]{article}

\usepackage[]{EMNLP2023}

\usepackage{color, colortbl}
\usepackage{times}
\usepackage{latexsym}
\usepackage{multirow}
\usepackage{booktabs}
\usepackage{makecell}
\usepackage{amssymb}
\usepackage{amsmath}
\usepackage{pifont}
\usepackage{graphicx}
\usepackage{tabularx}
\usepackage{bbm}
\usepackage{bm}
\usepackage{arydshln}
\usepackage{xcolor}
\usepackage{xspace}

\graphicspath{{./figures/}}
\usepackage[T1]{fontenc}
\usepackage[utf8]{inputenc}
\usepackage{microtype}
\usepackage{inconsolata}
\usepackage{booktabs}

\title{\toolkit: Lightweight Multilingual Development and Evaluation of Task-Oriented Dialogue Systems with Large Language Models}

\newcommand{\dataset}{\textsc{Multi3WoZ}\xspace}
\definecolor{ourred}{HTML}{E13342}
\definecolor{ourblue}{HTML}{6495ed}

\newcommand{\rparagraph}[1]{\vspace{1.2mm}\noindent\textbf{#1.}}
\newcommand{\rparagraphnodot}[1]{\vspace{1.2mm}\noindent\textbf{#1}}

\newcommand{\toolkit}{{\textsc{DiaLight}}\xspace}

\definecolor{Gray}{gray}{0.92}
\definecolor{racing-green}{rgb}{0.0, 0.8, 0.6}
\definecolor{awesome-red}{rgb}{1.0, 0.13, 0.32}
\newcolumntype{Y}{>{\centering\arraybackslash}X}

\usepackage{todonotes}
\makeatletter
\newcommand*\iftodonotes{\if@todonotes@disabled\expandafter\@secondoftwo\else\expandafter\@firstoftwo\fi}
\makeatother

\newcommand{\bluecheck}{{\color{racing-green}{\pmb{\checkmark}}}}
\newcommand{\xmark}{\color{awesome-red}\ding{55}}%

\usepackage[inline]{enumitem}

\newcommand{\tod}{{\textsc{ToD}}\xspace}

\usepackage{booktabs}
\usepackage{multirow}

\usepackage{float}

\usepackage{subfig}

\usepackage[]{pdfpages}

\usepackage[inline]{enumitem}

\author{
  Songbo Hu$^{1}$~~~ 
  Xiaobin Wang$^{2}$~~~ 
  Zhangdie Yuan$^{3}$~~~ 
  Anna Korhonen$^{1}$~~~
  Ivan Vuli\'{c}$^{1}$
  \\
  $^{1}$Language Technology Lab, University of Cambridge, UK
  \\
  $^{2}$Independent Researcher
  \\
  $^{3}$Department of Computer Science and Technology, University of Cambridge, UK
  \\
  $^{1, 3}$\texttt{\{sh2091,zy317,alk23,iv250\}@cam.ac.uk} \\
  $^{2}$\texttt{wxb9585@gmail.com} \\
}

\begin{document}
\maketitle

\begin{abstract}
We present \toolkit, a toolkit for developing and evaluating multilingual Task-Oriented Dialogue (\tod) systems which facilitates systematic evaluations and comparisons between \tod systems using fine-tuning of Pretrained Language Models (PLMs) and those utilising the zero-shot and in-context learning capabilities of Large Language Models (LLMs). In addition to automatic evaluation, this toolkit features (i) a secure, user-friendly web interface for fine-grained human evaluation at both local utterance level and global dialogue level, and (ii) a microservice-based backend, improving efficiency and scalability. Our evaluations reveal that while PLM fine-tuning leads to higher accuracy and coherence, LLM-based systems excel in producing diverse and likeable responses. However, we also identify significant challenges of LLMs in adherence to task-specific instructions and generating outputs in multiple languages, highlighting areas for future research. We hope this open-sourced toolkit will serve as a valuable resource for researchers aiming to develop and properly evaluate multilingual \tod systems and will lower, currently still high, entry barriers in the field.

\end{abstract}

\section{Introduction}
\label{sec:introduction}

Task-oriented dialogue (\tod) systems are designed to model interactions between human users and system agents, focusing on accomplishing specific, predefined tasks such as assisting with hotel or restaurant bookings, or providing domain-specific FAQ information~\cite{Gupta:2006, Tur:2010, Young:2010}. These systems serve not only as access points to cutting-edge AI applications but also as drivers of technological expansion.

The prevailing approach in \tod system development has predominantly involved fine-tuning Pre-trained Language Models (PLMs), like T5~\cite{t52020} and BART~\cite{lewis-etal-2020-bart}, on task-specific dialogue datasets~\cite{budzianowski-etal-2018-multiwoz, byrne-etal-2019-taskmaster}. However, recent research trends indicate a paradigm shift from fine-tuning to an increased reliance on Large Language Models' (LLMs) inherent capacity for in-context learning and  generalisation for natural language understanding and generation.
In our work, we categorise systems as fine-tuned-based using PLMs (hereafter FT-based) or in-context-learning-based using LLMs (ICL-based), noting that LLMs can be fine-tuned and smaller PLMs can use ICL.\footnote{\textbf{Disclaimer:} Here, we (coarsely) differentiate between PLMs and LLMs in terms of their dependency on fine-tuning with task-specific datasets for achieving optimal performance. LLMs, such as LLaMA~\cite{touvron2023llama} and GPT-4~\cite{OpenAI2023GPT4TR}, are characterised by their extensive training on a broad spectrum of data. This approach enables LLMs to adapt to diverse tasks  with minimal reliance on task-specific data. Empirical evidence suggests that LLMs demonstrate remarkable capabilities, nearing or `surpassing' human-level performance on NLP benchmarks such as SuperGLUE~\cite{superglue} and BIG-Bench~\cite{srivastava2023beyond}.}

Several pilot works~\cite {hudecek-dusek-2023-large, heck2023chatgpt, zhang2023sgp, chung2023instructtods} have explored the integration of LLMs into \tod systems. These studies indicate that FT-based approaches outperform ICL-based approaches, as evidenced by superior automatic evaluation scores. This applies even with smaller PLMs and when the number of training examples is limited. On the other hand, instruction-based training of LLMs demonstrates its potential in aligning model outputs more closely with human preferences~\cite{ouyang2022training, wang2022super}. In \tod, ICL-based systems~\cite{chung2023instructtods} have been shown to generate responses that exceed previous models in critical human evaluation dimensions such as informativeness, helpfulness, and perceived humanness. Despite these early results, a systematic, comparative human evaluation of these two approaches in \tod systems remains a gap in current research.\footnote{The study by~\citet{chung2023instructtods} lacks full details on their human evaluation protocol, leading to potential ambiguities in interpretation. Moreover, current research infrastructure falls short in facilitating extensive human evaluation of end-to-end \tod systems, especially for comparative analyses between FT-based and ICL-based systems.}

Furthermore, the development of \tod systems has historically been confined to a limited number of high-resourced languages~\cite{Razumovskaia:2022survey}. The recent release of the Multi3WOZ dataset~\cite{multi3woz} expands the linguistic scope, introducing the same-level data support for Arabic, English, French, and Turkish \tod. Nevertheless, there are still noticeable disparities in system performance across different languages even with the fully comparable training data~\cite{hu2023systematic}, raising questions about system utility and user satisfaction with \tod in non-English contexts. To facilitate future research in minimising these performance disparities, a toolkit that supports developing and evaluating multilingual dialogue systems is critically needed.

Aiming to address these gaps, this paper introduces \toolkit, a novel toolkit for developing and evaluating multilingual end-to-end (E2E) \tod systems. \toolkit is specifically designed for comprehensive comparative analyses between FT-based and ICL-based systems (see \S \ref{sec:system}). It supports an array of \tod datasets from the MultiWOZ family~\cite[\textit{inter alia}]{Budzianowski:2018multiwoz, ding-etal-2022-globalwoz, multi3woz}, and enables seamless evaluations in monolingual, multilingual, and cross-lingual setups (\S \ref{sec:auto_eval}). Given often the moderate correlation between automatic evaluation metrics and human judgements~\cite{yeh-etal-2021-comprehensive, mehri2022report}, our toolkit places a special emphasis on human evaluation, facilitating both utterance-level and full dialogue-level assessments (\S \ref{sec:human_eval}). The toolkit is available at \url{https://github.com/cambridgeltl/e2e_tod_toolkit}.

\section{Related Work}
\label{sec:rw}
\begin{table*}[!t]
\centering
\def\arraystretch{0.9}
\footnotesize
\resizebox{\textwidth}{!}{%
\begin{tabular}{lcccc}
\toprule
\rowcolor{Gray}
\textbf{Toolkit}                                    & \multicolumn{1}{l}{\bf Human Evaluation} & \multicolumn{1}{l}{\bf Multilingualilty} & \multicolumn{1}{l}{\bf LLM+E2E} & \multicolumn{1}{l}{\bf Comparative Experiment} \\ \midrule
PyDial~\cite{ultes2017pydial}              & \bluecheck                           & \xmark                           & \xmark                      & \xmark                                     \\
ConvLab2~\cite{zhu2020convlab2}            & \bluecheck                           & \xmark                           & \xmark                      & \xmark                                     \\
ConvLab3~\cite{zhu2022convlab3}            & \bluecheck                           & \bluecheck                       & \xmark                      & \xmark                                     \\
to-llm-bot~\cite{hudecek-dusek-2023-large} & \xmark                               & \xmark                           & \bluecheck                  & \xmark                                     \\
other E2E baselines                        & \hspace{3mm} \xmark {\color{black}$(*)$}                          & \hspace{3mm} \xmark {\color{black}$(*)$}                      & \xmark                      & \xmark                                     \\
\midrule
\toolkit (this work)                                      & \bluecheck                           & \bluecheck                       & \bluecheck                  & \bluecheck                                 \\ \bottomrule
\end{tabular}%
}
\caption{A comparative overview of \tod system toolkits supporting E2E modelling. This summary excludes system components focused solely on dialogue state tracking (DST) and response generation (RG). Key features include \textbf{Human Evaluation}, indicating support for online crowdsourcing human evaluation, \textbf{Multilingualilty} capabilities for development and evaluation across various languages and in monolingual, multilingual, and cross-lingual setups, \textbf{LLM+E2E} for in-context learning with LLMs in E2E modelling, and \textbf{Comparative Experiment}, denoting a unified framework for evaluating both FT-based and ICL-based systems. Detailed comparisons are presented in Section~\ref{sec:rw}. $(*)$ We acknowledge that while some \tod systems support human evaluation and multilingualism, it is recognised that a significant majority lack these crucial features.}
\label{tab:rw}
\end{table*}

\toolkit represents a novel addition to the landscape of \tod system toolkits, complementing existing frameworks such as PyDial~\cite{ultes2017pydial}, ConvLab-3~\cite{zhu2022convlab3}, and their antecedents~\cite{zhu2020convlab2, lee-etal-2019-convlab}, as shown in Table~\ref{tab:rw}. \toolkit is unique in offering support for ICL-based implementations in E2E ToD systems, a feature not yet available in existing toolkits.
Moreover, \toolkit diverges in its core design philosophy. Instead of incorporating extensive and intricate systems and components, it is meticulously crafted to reduce entry barrier and learning curve for researchers engaging in multilingual \tod research.\footnote{The term \textit{`reduce the learning curve'} in this context comes with positive but subjective connotations. \toolkit is designed for easier use with fewer code lines and a simplified interface. It aims to lower the entry barrier in multilingual \tod research. However, we acknowledge that individual factors like prior experience, technical background, and learning styles influence the learning curve.}
Our objective is to provide a streamlined codebase that facilitates rapid development of multilingual \tod systems.

Another range of publicly available implementations exists for both traditional FT-based and ICL-based E2E \tod systems, typically accompanying research publications as supplementary code. For traditional FT-based approaches, works include those by~\citet{wen2016network, bordes2017learning, lei2018sequicity, eric-manning-2017-copy, eric-etal-2017-key, lin2020mintl, peng-etal-2021-soloist, he2022galaxy} . The ICL-based category is exemplified by the work of~\citet{hudecek-dusek-2023-large}.\footnote{To date (December 2023), the implementations by~\citet{zhang2023sgp, chung2023instructtods} have not been released.} 
However, these systems lack a unified setup for comparative experimentation. Our toolkit fills this gap, enabling fair and comprehensive comparisons between the aforementioned two types of systems. Additionally, many of these systems, primarily designed to enhance benchmark results, exhibit key limitations, such as:
\begin{enumerate*}
    \item absence of implementation for lexicalisation in utterances,
    \item lack of a dialogue system agent for real user interaction,
    \item existence of English-centric heuristics in evaluations,\footnote{For instance, the official evaluation script for MultiWOZ~\cite{nekvinda-dusek-2021-shades} employs a string matching algorithm that normalises English slot values to their canonical forms. This approach, however, introduces a bias in evaluations, leading to potentially unfair comparisons when extending the framework to other languages.}
    \item inadequate support for human evaluation.
\end{enumerate*}
\toolkit specifically addresses these limitations.

While existing tools for human evaluation, such as DialCrowd~\cite{lee-etal-2018-dialcrowd, huynh-etal-2022-dialcrowd}, offer simple approaches for conducting human evaluation experiments in \tod systems, they are not without limitations.
DialCrowd is not open-sourced, available solely via its designated website, which imposes significant constraints.
These include limited customisation flexibility, increased maintenance complexity,\footnote{E.g., DialCrowd is currently offline (December 2023).} and challenges in aligning with data protection regulations such as GDPR, thus affecting its broader applicability in research. In contrast, our human evaluation tool is open-sourced, enabling `one-click' deployment on local or cloud servers.

\section{System Architecture}
\label{sec:system}
\begin{figure}[!t]
    \centering
    \includegraphics[width=0.9\linewidth]{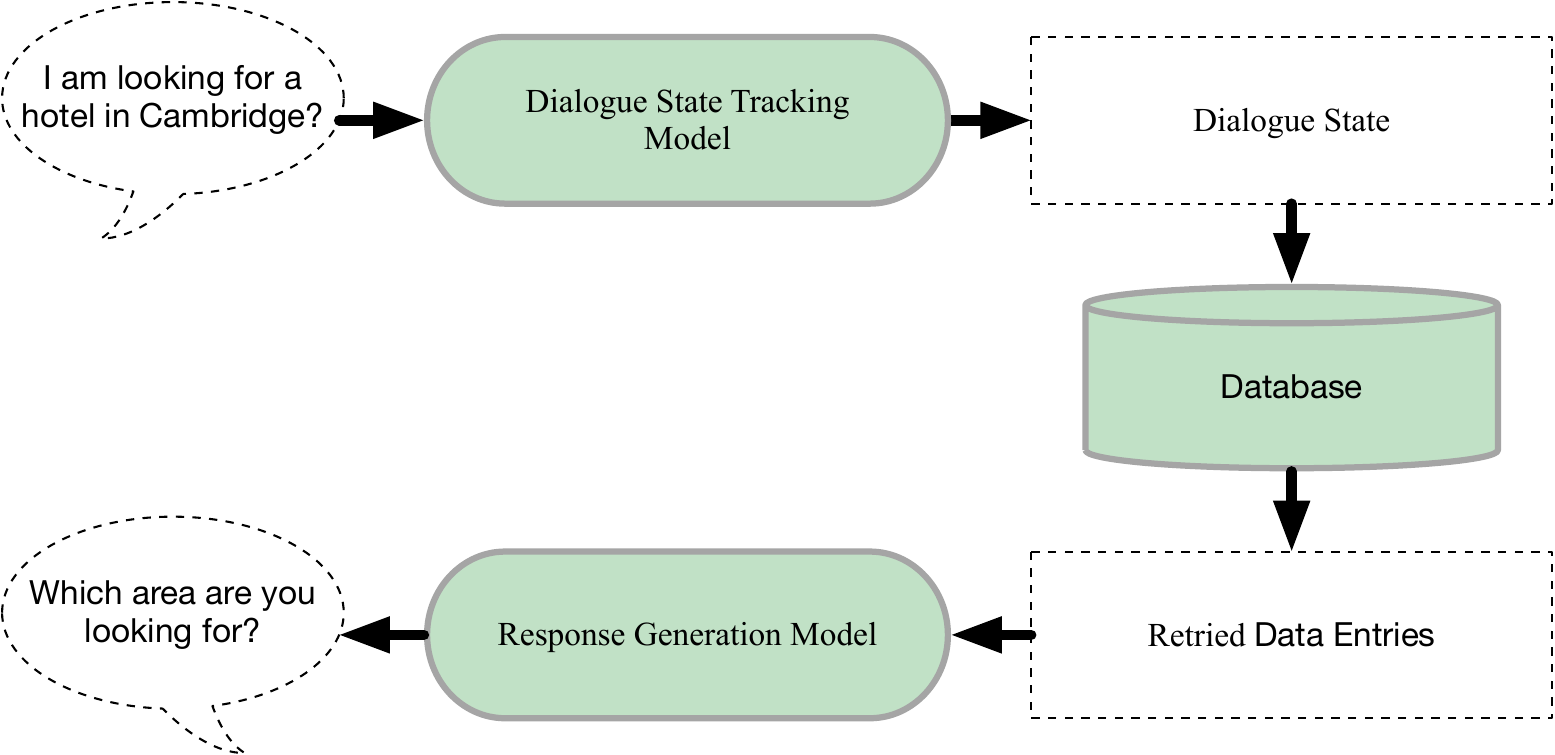}
    \caption{An E2E dialogue system contains three key components: a dialogue state tracking (DST) model, a database, and a response generation (RG) model. The DST model processes user utterances with the accumulated dialogue history to predict a dialogue state. This state is then translated into a database query to extract data entries relevant to the current dialogue context from the database. The RG model uses these entries and the dialogue history to produce the final response.}
    \label{fig:pipeline}
\end{figure}

This section delineates the architecture and implementation of the proposed E2E dialogue system within \toolkit. As shown in Figure~\ref{fig:pipeline}, despite the term `end-to-end', state-of-the-art E2E \tod systems typically employ a pipelined approach in the background, incorporating three key components: a dialogue state tracking (DST) model, a database interface, and a response generation (RG) model. In the following, we will explain the pipeline of our system and provide an in-depth examination of each constituent component.

\subsection{System Pipeline}

Our E2E  \tod system operates by processing a dialogue history, represented as the concatenation of a list of preceding dialogue utterances $[\mathbf{u}_{1}, \mathbf{u}_{2}, \cdots, \mathbf{u}_{t-1}]$ with the latest user utterance $\mathbf{u}_{t}$. Specifically, the DST model, denoted as $\operatorname{DST}(\cdot)$, takes this concatenated input dialogue utterances to predict the current dialogue state, formulated as $\mathbf{s}_{t} = \operatorname{DST}([\mathbf{u}_{1}, \mathbf{u}_{2}, \cdots, \mathbf{u}_{t}])$. Within the context of MultiWOZ datasets, a dialogue state is defined as a set of tuples $\mathbf{s} = \{(d_1, s_1, v_1), \cdots, (d_k, s_k, v_k)\}$, where each tuple consists of domain $d$, slot $s$, and slot value $v$. 
Then, this state is sent to a database $\operatorname{DB}(\cdot)$, from which a set of entities satisfying the requirements specified by the state $\mathbf{s}_{t}$ are retrieved. Namely,  $\{\mathcal{E}_1, \cdots \mathcal{E}_{l}\} = \operatorname{DB}(\mathbf{s}_{t})$, where $\mathcal{E}$ is a data entry in the database.
The RG model, denoted as $\operatorname{RG}(\cdot)$, is then tasked to consume the sequence of dialogue utterances with the retrieved set of entities as input and generate a dialogue response $\mathbf{u}_{t+1}$. The process can be formally expressed as $\mathbf{u}_{t+1} = \operatorname{RG}([\mathbf{u}_{1}, \cdots, \mathbf{u}_{t}], \{\mathcal{E}_1, \cdots, \mathcal{E}_{l}\})$, where $\mathbf{u}_{t+1}$ represents the generated dialogue response to user input $\mathbf{u}_{t}$, taken as the next system turn.

In the proposed toolkit, DST models and RG models can be implemented using both FT-based and ICL-based methods. In what follows, we detail the implementation for each system component, applying both methods.

\subsection{Dialogue State Tracking Models}
\rparagraph{Fine-Tuning with PLMs}
The dialogue state $\mathbf{s}_{t}$ is transformed into a flattened string representation. For example, consider a dialogue state $\{(d_1, s_1, v_1), \cdots, (d_k, s_k, v_k)\}$, where $d_1$=$d_2$, the state is transformed it into a flattened string representation of $d_1$\#$s_1$=$v_1$;$s_2$=$v_2$|  $\cdots$|$d_k$\#$s_k$=$v_k$. In this representation, slots and their corresponding values within the same domain are merged. For example, the dialogue state `\{(taxi, departure, saint johns college), (taxi, destination, pizza hut fenditton)\}' is linearised as `taxi \# departure = saint johns college ; destination = pizza hut fenditton'. At each dialogue turn $t$, a PLM is trained to take the input of the concatenated dialogue history $[\mathbf{u}_{1}, \mathbf{u}_{2}, \cdots, \mathbf{u}_{t}]$, and generates the linearised dialogue state.

\rparagraphnodot{In-Context Learning with LLMs.}
We use LLMs for direct generation of dialogue states in JSON format, guided by task-specific prompts and training examples for in-context learning. The instruction prompt consists of six parts:
\begin{enumerate*}
\item \textit{Task instruction} for generating the dialogue state.
\item \textit{Output format instruction} to produce results in JSON format.
\item \textit{Ontology instruction} detailing available domains and slots.
\item \textit{Categorical slot instruction}, guiding LLMs to choose from predefined options for categorical slots.
\item \textit{Time slot instruction} for generating times in 24-hour format \texttt{(hh:mm)}.
\item \textit{Number slot instruction}, directing LLMs to produce non-negative integer values for numeric slots, such as the quantity of individuals in a booking.
\end{enumerate*}
After these instructions, we incorporate a list of training examples, randomly chosen from the training dataset, to provide a baseline for future research. The final step involves appending the concatenated dialogue history $[\mathbf{u}_{1}, \mathbf{u}_{2}, \cdots, \mathbf{u}_{t}]$ to the prompt, enabling the LLMs to generate the corresponding dialogue state $\mathbf{s}_{t}$.

\rparagraph{Implementation and Setup}
In \toolkit, the FT-based DST models can be instantiated with any of the PLMs available in the Huggingface repository~\cite{wolf2019huggingface}. Additionally, we provide comprehensive support for various models in ICL-based systems, including:
\begin{enumerate*}
\item Models from the Huggingface repository,
\item Models accessible through the OpenAI API,\footnote{\url{https://openai.com/blog/openai-api}}
\item LLaMA.cpp models, tailored for on-device LLM inferences.\footnote{LLaMA.cpp is open-sourced and accessible at \url{https://github.com/ggerganov/llama.cpp}.}
\end{enumerate*}
For evaluation of FT-based DST in this paper, we utilise the mT5\textsubscript{small} and mT5\textsubscript{large}~\cite{xue-etal-2021-mt5}. For ICL-based DST experiments, we employ the GPT-3.5, LLaMA2~\cite{touvron2023llama}, and OpenChat-3.5~\cite{wang2023openchat} models.\footnote{All the other experimental details are in Appendix~\ref{sec:experiment_details}.}

\subsection{Database Interface}
We adapt the implementation of our database interface from the official MultiWOZ evaluation scripts~\cite{nekvinda-dusek-2021-shades} with minor modifications. 
Each data entity across different domains within the database can be represented as a set of slot-value pairs.\footnote{As an illustration, consider the following example of a database entry in the \textit{police} domain: \textit{\{(name, Parkside Police Station), (address, Parkside Cambridge), (phone, 01223358966), (postcode, CB11JG)\}}.
}
In response to each user utterance $\mathbf{u}_{t}$, our system executes a database query using the predicted dialogue state $\mathbf{s}_{t}$, to independently retrieve relevant data entries from each respective domain.
For each domain, a database entry is retrieved if it meets the following criteria:
\begin{enumerate*}
\item exact matching of categorical slot values with those prescribed in the dialogue state,
\item achieving a Levenshtein distance for non-categorical slot values that is below a predefined threshold when compared with the dialogue state.
\end{enumerate*}\footnote{A categorical slot is defined by the ontology such that the possible values for this slot are a closed set. For example, the slot `price range' can only have the values of `cheap', `moderate', and `expensive'. In contrast, the value for a \textit{hotel name} is an open set and not categorical, as it can be any string.}

\subsection{Response Generation Models}

\rparagraph{Fine-Tuning with PLMs}
In the context of the MultiWOZ datasets, our approach follows the conventional two-step process: initially generating a delexicalised response and subsequently lexicalising this response with slot values from dialogue states and retrieved entities.\footnote{In the case of delexicalised dialogues, all the slot values in the context and responses are replaced a predefined placeholder (e.g. \textit{[value\_name] is an [value\_price] [value\_food] restaurant on the [value\_area] . do you need to know more ?}).}
It is worth noting that the majority of prevalent automatic evaluation metrics predominantly focus on delexicalised responses. However, incorporating lexicalisation is a more realistic scenario, which is also a necessity for proper human evaluation.

The generation of a delexicalised response is modelled as a transduction problem, converting dialogue history into a natural response.
To integrate database outcomes, we initially create a summary (e.g., \textit{`attraction has one result found; hotel has no result found'}). Subsequently, a PLM is trained to process the input, which is a combination of dialogue history and the database summary, to produce the delexicalised system response. The lexicalisation process is carried out through a systematic replacement of placeholders with the relevant values from the current dialogue state $\mathbf{s}_{t}$ and from the retrieved entities.

\rparagraphnodot{In-Context Learning with LLMs.}
The ICL-based method also follows the conventional two-step process. However, instead of fine-tuning, this approach involves prompting LLMs with a task-specific instruction that encompasses four parts:
\begin{enumerate*}
\item \textit{Task instruction}, which directs the generation of the dialogue response.
\item \textit{Ontology instruction}, providing details on the available domains and slots.
\item \textit{Delexicalisation instruction}, informing the LLM about all available placeholders and guiding it to substitute slot values with these placeholders.
\item \textit{Language instruction}, specifying the target language for the generated response.
\end{enumerate*}
A set of training examples, randomly selected from the dataset, is appended to these instructions. The LLM is then tasked to generate the corresponding dialogue response based on this augmented propmt, the database summary, and the concatenated dialogue history $[\mathbf{u}_{1}, \mathbf{u}_{2}, \cdots, \mathbf{u}_{t}]$ .

\rparagraphnodot{Implementation and Setup.}
Similarly as before with DST models, we employ the mT5\textsubscript{small} and mT5\textsubscript{large} models for FT-based response generation. For ICL-based experiments, we employ the GPT-3.5, LLaMA2, and OpenChat-3.5 models.

\section{Automatic Evaluation}
\label{sec:auto_eval}

\begin{table*}[!t]
\centering
\def\arraystretch{0.75}
\fontsize{6.8}{7.2}\selectfont
\resizebox{0.91\textwidth}{!}{%
\begin{tabular}{@{}lccccclccclccc@{}}
\toprule
                           &                      & \multicolumn{4}{c}{\textbf{Dialogue State Tracking}}                                & \multicolumn{1}{c}{} & \multicolumn{3}{c}{\textbf{Response Generation}} &  & \multicolumn{3}{c}{\textbf{End-to-end Modelling}} \\ \cmidrule(lr){3-6} \cmidrule(lr){8-10} \cmidrule(l){12-14} 
\multirow{-2}{*}{Language} &                      & JGA                  & Slot F1              & Slot Precision          & Slot Recall &                      & BLEU          & ROUGE          & METEOR          &  & Inform Rate       & Success Rate      & BLEU      \\ \midrule
\multicolumn{14}{c}{\cellcolor[HTML]{EFEFEF}FT-mT5\textsubscript{small}}                                                                                                                                                                                                 \\ \midrule
ENG                     &    &   56.4                &     83.4                 &           83.7           &        83.1                   &                      &    16.3           &  24.5              &          27.2       &  &     70.7          &        46.0        &  16.3        \\
ARA                        &                      &                43.8                &     77.5                 &           78.7           &        76.3            &                      &      15.0      &      28.8      &       26.4      &  &        70.2         &     42.6          &      15.0     \\
FRA                        &                      &                    47.0                &     79.6                 &           79.5           &        79.8             &                      &     14.3        &     25.5       &       26.3        &  &    71.5        &            41.5   &    14.1   \\
TUR                        &                      &       48.9                &     80.7                 &           78.8           &        82.6               &                      &    21.0       &   33.7        &       33.3  &  &       77.7        &       48.0        &    21.4    \\
AVG.                       &                      &        49.0      &   80.3               &       80.2         &      80.5      &                      &     16.6        &  28.1       &        28.3       &  &       72.4      &     44.4         &  16.7   \\ \midrule
\multicolumn{14}{c}{\cellcolor[HTML]{EFEFEF}FT-mT5\textsubscript{large}}                                                                                                                                                                                                 \\ \midrule
ENG                        &                      &    18.6       &      52.5           &    53.0           &    52.0       &                      &      15.8     &      24.1          &       27.1          &  &       70.1       &           47.3        &      15.4     \\
ARA                        &  &  44.0 &  78.5 & 78.3 &  78.6              &                      &        7.0       &   17.4             &     14.8            &  &  67.3                &    31.4            &  6.7          \\
FRA                        &                      &        46.8              &           79.5           &             79.6         &       79.5         &       &      14.1              &       25.2        &     25.8                         &  &               74.3         &   44.2          &     13.6      \\
TUR                        &             &      48.5                    &      80.2          &     78.8                 &     81.7       &                      &       10.6      &     19.7           &        19.0         &  &         77.7     &    48.0             &      10.8 \\
AVG.                       &                      &   39.5             &      72.7              &      72.4               &   73.0           &                      &      11.9         &       21.6         &      21.7           &  &   70.1         &   42.4          &     11.6    \\ \midrule
\multicolumn{14}{c}{\cellcolor[HTML]{EFEFEF}{\color[HTML]{000000} ICL-GPT-3.5 {\color{black}$(*)$}}}                                                                                                                                                                                            \\ \midrule
ENG                        &   &  13.5 & 37.6 & 26.3 &       65.8         &                      &      1.8         &       12.2         &   11.4              &  &       33.0         &       16.0         &    1.8       \\
ARA                        &  &  7.6 & 31.9 & 22.6 &         60.2       &                      &     0.1       &        2.2       &      1.5           &  &         36.0        &        18.0           &     0.1      \\
FRA                        &   & 12.5 & 39.4 & 29.0 &       61.2         &                      &     0.9          &      8.7          &    7.2             &  &        40.0         &       24.0            &   0.8        \\
TUR                        &   &   7.7  &  34.5  & 24.7 &   57.4          &                      &     0.8          &    6.1            &   5.0              &  &             32.0      &      13.0             &     0.8      \\
AVG.                       &   & 10.3 & 35.9 &   25.7             &        61.2             &   &      1.2         &        7.3        &      6.3           &  &     35.3              &    17.8         &   0.9        \\ 

\bottomrule
\end{tabular}%
}
\caption{Evaluation of fully supervised performance across DST models, RG models, and E2E systems on the \dataset dataset. This table reports the performance metrics for each language, evaluated across different models. The notation FT-mT5\textsubscript{small} indicates that the respective system component, or the entire system, is backboned with the fine-tuned mT5\textsubscript{small} model. 
Furthermore, ICL-GPT-3.5 refers to the models or systems developed based on in-context learning with the GPT-3.5 model. `AVG.' represents the mean average of the evaluation scores aggregated across all four languages. We note that for these metrics the ground truth score is set at 100, with the exception of the Inform Rate and Success Rate, which are measured as $89.3\pm0.2$ and $68.6\pm0.2$ across the four languages, respectively. $(*)$ For practical considerations, the evaluation of ChatGPT-3.5-based models and systems is limited to a randomly selected sample of 100 dialogues from the full test set, due to the significant time and resource requirements of full-scale evaluation.}
\label{tab:auto_eval}
\end{table*}

As part of \toolkit, we have implemented a range of automatic evaluation metrics:
\begin{enumerate*}
\item For DST evaluation, metrics include Joint Goal Accuracy (JGA),\footnote{The JGA measure represents the proportion of dialogue turns in the dataset where all slots have been correctly filled with their ground truth values.} Slot F1, Slot Recall, and Slot Precision.
\item For response generation evaluation, we utilize the BLEU score~\cite{papineni2002bleu}, ROUGE-L~\cite{lin-2004-rouge}, and METEOR~\cite{banerjee2005meteor}.
\item For evaluating the overall systems, we report Inform Rate, Success Rate, and BLEU.
\end{enumerate*}
Additionally, we provide an interface to facilitate easy extension for future additional metrics. Table~\ref{tab:auto_eval} presents the performance of our systems across various languages and backbone models, evaluated using the aforementioned automatic metrics.\footnote{We show the performance of the ICL-LLaMA2 and ICL-OpenChat-3.5 systems in Table~\ref{tab:auto_eval_appendix} in the Appendix. Additionally, Table~\ref{tab:auto_eval_100} in the Appendix presents the evaluation results of FT-based systems, specifically on the same selected subset of 100 dialogues for consistency in comparison.} The main results indicate a performance advantage of FT-based systems over their ICL-based counterparts.  Furthermore, the evaluation metrics for E2E tasks, such as Inform Rate and Success Rate, are influenced by two key outputs of the system: the dialogue state and the generated response. To assess the impact of each component on overall system performance, we conduct an extra experiment where the predictions of each part were individually replaced with the ground-truth. When substituting the predicted utterances with ground-truth utterances, the FT-mT5\textsubscript{small} systems exhibited a notable improvement, achieving an average Inform Rate of 85.1 ($\uparrow$13.3) and a Success Rate of 66.1 ($\uparrow$21.5) across four languages. In contrast, substituting the predicted dialogue states resulted in a marginal increase, with the systems attaining an Inform Rate of 72.1 ($\uparrow$0.3) and a Success Rate of 43.3 ($\uparrow$0.7) across the languages. These findings highlight the critical role of RG model performance in determining overall system performance and the relative insensitivity of these metrics in evaluating the performance of DST models.

The ICL-GPT-3.5-based \tod system demonstrates inferior performance when compared to the FT-based systems, as shown in Table~\ref{tab:auto_eval}.\footnote{It is worth noting that our system's simplicity, as opposed to the more complex system proposed by~\citet{hudecek-dusek-2023-large}, leads to its lower absolute performance. Our system is designed to serve as a baseline for future research.} Subsequently, we present a detailed analysis aimed at identifying the root causes of this performance discrepancy. Initially, we observe that a substantial portion, 42.7\%, of the system predictions generated by the English ICL-based DST model do not adhere to the prescribed dialogue state format specified by the instruction and ontology. Furthermore, with the given instructions, the ICL-GPT-3.5 system generates delexicalised English system responses that recall only 3.6\% of the placeholders found in the ground-truth utterances. Concerning other languages, even when explicitly instructed to generate responses in the target language, the GPT-3.5 model produces utterances thare are only 18.4\% in Arabic, 78.0\% in French, and 70.5\% in Turkish.\footnote{Language detection was performed using the tool developed by~\citet{nakatani2010langdetect}.} We hypothesise that this is attributed to the fact that our prompt is provided in English, and future work should experiment with additional and more sophisticated prompt designs.

\section{Human Evaluation}
\label{sec:human_eval}

In \toolkit, we offer an open-sourced human evaluation tool specifically designed for \tod systems. This tool is comprehensive, providing all essential functionalities to conducting human evaluation experiments in a production setting. In the following, we detail the features supported by our web interface, provide an overview of the high-level design of our backend servers, and present a case study of human evaluation experiments.

\subsection{Web Interface}
The web interface of the tool offers a range of features, including user registration, account login, consent acquisition for data collection, and the execution of human evaluation tasks. In this section, we highlight two critical features that distinguish our tool from existing work.

\begin{figure}[!t]
    \centering
    \subfloat[Utterance Level]{\includegraphics[width=0.85\linewidth]{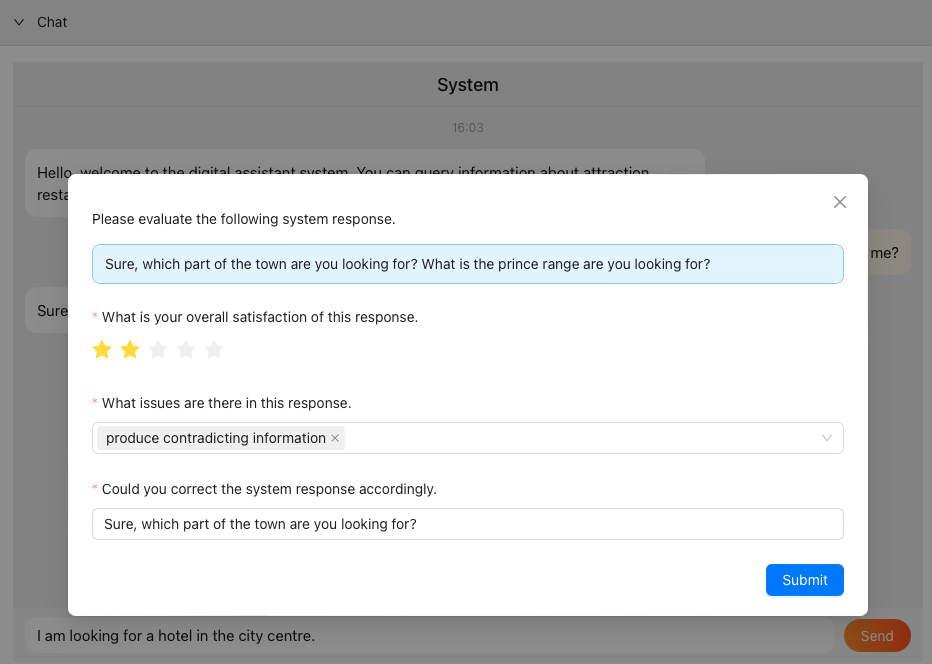}}%
    \quad
    \subfloat[Dialogue Level]{\includegraphics[width=0.85\columnwidth]{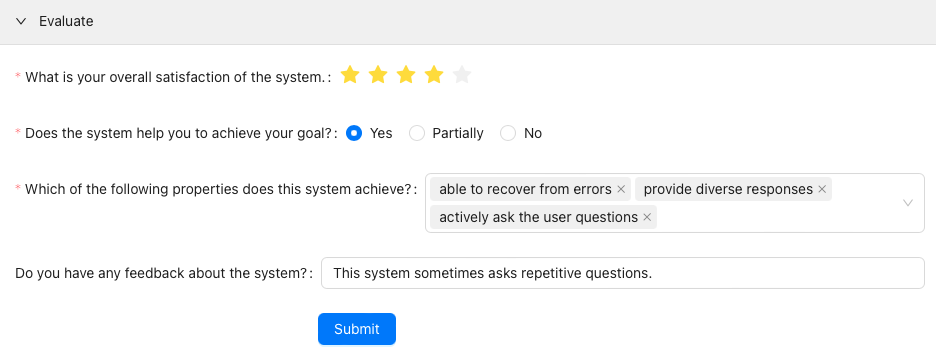}}%
    \vspace{-1mm}    
    \caption{Our human evaluation tool is designed to collect user feedback at both the (a) utterance and (b) dialogue levels. This tool allows for full customisation of evaluation questions with minimal programming effort.}%
    \label{fig:feedback_screenshot}%
    \vspace{-1.5mm}
\end{figure}

\rparagraphnodot{Fine-Grained User Feedback.}
As shown in Figure~\ref{fig:feedback_screenshot}, our web interface is designed to support the collection of user feedback and scores at both the (local) utterance and (global) dialogue levels. Furthermore, all evaluation questions can be fully customised with minimal programming effort.\footnote{Our tool supports all components provided by the AntDesign toolkit, available at \url{https://ant.design/components/overview}.}

\rparagraphnodot{Secure Authentication and Access Management.}
A high priority has been placed on data security and the incorporation of authentication measures. An authentication system employing JSON Web Tokens (JWT) and a basic role management framework has been implemented. This arrangement ensures that access to specific task groups is limited to authorised users, and only task administrators are permitted to access submitted data in the database. Moreover, the web interface is integrated with an \texttt{nginx} reverse proxy, enhancing data and communication security through SSL encryption.\footnote{These measures align with GDPR requirements for Data Protection by Design and by Default.}

\subsection{Back-End Servers}

\begin{figure}[!t]
    \centering
    \includegraphics[width=0.75\linewidth]{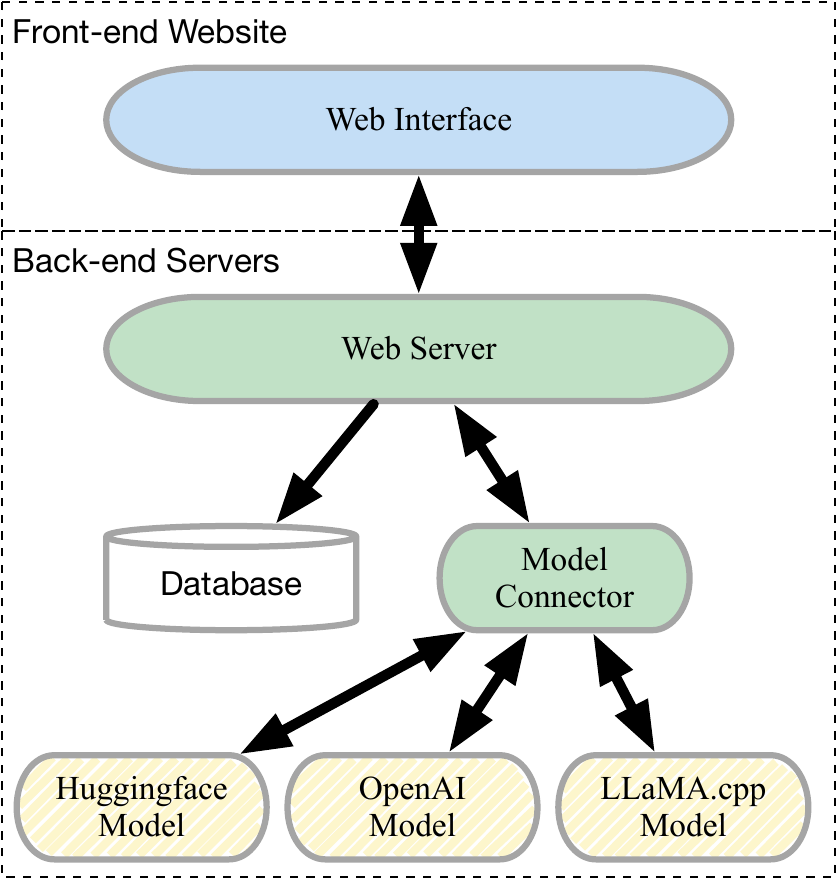}
    \caption{Architectural design of our human evaluation tool, containing two primary components: a web-based interface and a cluster of back-end servers. The server infrastructure is based on a microservice architecture, segregating each task model into its own independent service, as highlighted by the dashed yellow background in the figure. Central to this architecture is the Model Connector, functioning as an API gateway to manage and route requests to the appropriate task models. For example, each fine-tuned DST model is hosted independently on a dedicated server. These servers are designed to be stateless, enabling their shared use across various systems and dialogue sessions, thereby enhancing efficiency and scalability.}
    \label{fig:servers}
    \vspace{-1mm}
\end{figure}

Our human evaluation tool is architected using a microservice design, a choice that significantly enhances its scalability and adaptability. Another standout feature of the tool is the `one-click' deployment option, making it more accessible for users. This tool sets itself apart from its predecessors by being easy-to-use and tailored for the prevailing trend of LLMs.

\rparagraphnodot{Microservice with Scalability.}
In a microservice architecture, each service operates independently, managing a specific task or functionality. Our tool adopts this modular framework, partitioning each task model into its own independent service on a dedicated server, as depicted in Figure~\ref{fig:servers}. The design of these stateless services offers several benefits: firstly, it enables a single model to be concurrently shared by multiple systems, each with different configurations; secondly, it allows for the deployment of multiple instances of the same model within the same system.\footnote{This approach stands in stark contrast with previous \tod tools~\cite{ultes2017pydial, zhu2020convlab2, zhu2022convlab3}, which instantiate and load multiple task models (e.g., DST or response generation models) on the same device, a method that becomes increasingly restrictive and computationally demanding with the integration of LLMs. For systems utilising multiple LLM-based components, finding a host machine capable of supporting two or more LLMs simultaneously is non-trivial. Additionally, given that LLMs can be time-consuming in generating responses, our architecture facilitates load balancing by distributing a set of queries across multiple instances of the same task model distributed on different servers, thereby optimising large-scale experimental setups.}

\rparagraphnodot{On-click Deployment.}
The tool is designed with an `out-of-the-box' capability, facilitated by full containerisation using docker and docker-compose.\footnote{\url{https://docs.docker.com/}} This approach ensures a simple and efficient deployment process. The versatility of these containers supports deployment in various environments, ranging from local machines to cloud-based platforms. The entire build process is governed by a central configuration file, which users can modify according to their specific requirements, thus enhancing the tool's adaptability.

\subsection{A Pilot Analysis of FT-based versus ICL-based Systems in English}

\begin{figure}[!t]
    \centering
    \includegraphics[width=0.86\linewidth]{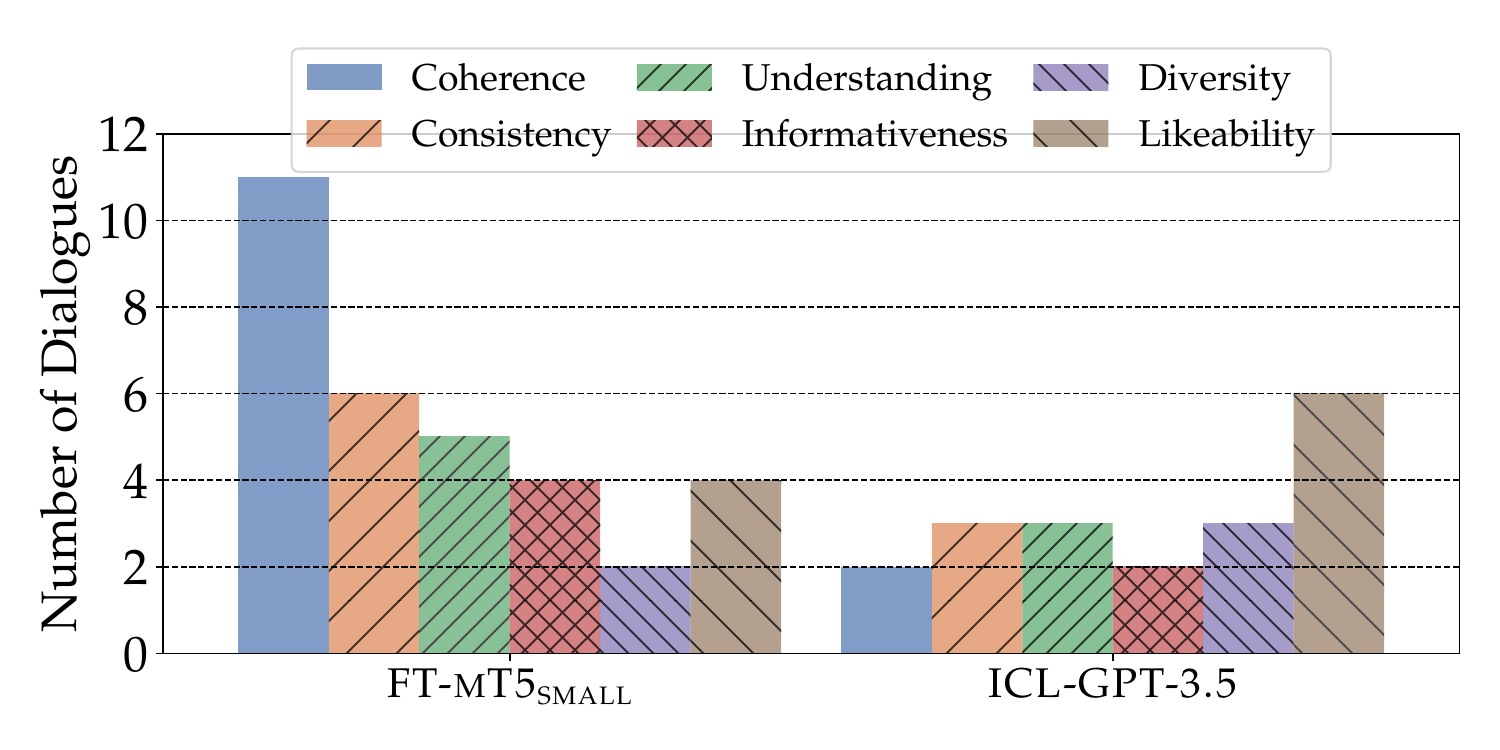}
    \vspace{-1.5mm}
    \caption{Number of dialogues (from the total of 20) assessed by human evaluators according to the desirable properties that align with the dialogue-level evaluation dimensions outlined in~\citet{mehri-eskenazi-2020-unsupervised}.}
    \label{fig:dial_rate}
    \vspace{-2mm}
\end{figure}

Utilising \toolkit, we conduct a human evaluation experiment comparing the performance of the English system with FT-mT5\textsubscript{small} and the system utilising the ICL-GPT-3.5 model. In this study, each of the 10 participants has completed 2 dialogues for each system, resulting in a total of 40 data entries. The FT-based system attains an overall score of $3.8 \pm 0.9$, while the ICL-based system achieves $1.6 \pm 0.9$. Furthermore, Figure~\ref{fig:dial_rate} demonstrates the number of dialogues for each system based on a set of dialogue-level evaluation dimensions~\cite{mehri-eskenazi-2020-unsupervised}. The results indicate that the FT-based system outperforms the ICL-based system in terms of maintaining conversation coherence, providing consistent information, understanding the user, and delivering informative responses. Conversely, the ICL-based system generates more diverse responses and exhibits a more favourable personality.

\section{Conclusion and Outlook}
\label{sec:conclusion}
We introduced \toolkit, a comprehensive toolkit for advancing multilingual \tod systems supported by language models of different families, offering the essential infrastructure for system development and evaluation. \toolkit supports the development of and includes baseline systems for fine-tuning and in-context learning paradigms, enabling comparative experiments within a unified setup for automatic and human evaluation. We have publicly released all our source code to facilitate future research and invite the research community to adapt and contribute to this toolkit.

Based on the developed toolkit, we conducted a performance analysis across several languages and models for the two modeling paradigms. The results reveal that systems developed through fine-tuning PLMs with in-domain in-language data outperform systems based on LLMs through ICL paradigms. Conversely, our human evaluation results indicate that LLMs generate responses that are more diverse and exhibit a greater degree of likeability. In summary, our results indicate that despite their potential, LLMs, even the most powerful ones, are far from `solving' the \tod task, especially in a multilingual context. Instead, our results open up new avenues for future research and exploration. In the following, we provide examples of future work aimed at addressing the bottlenecks identified in the baseline systems.

\rparagraph{Modernising \tod Systems for LLMs}
In \S\ref{sec:auto_eval}, our experimental results reveal that solely prompting LLMs resulted in system failure to comply with the instructions and predict dialogue states in the required format, leading to over 40\% empty predictions. Additionally, ICL-based systems employing LLMs encountered challenges in recalling placeholders and generating delexicalised utterances in a similar fashion. This issue is arguably attributable to the misalignment between the task requirements for \tod and the inherent pretraining of LLMs. We propose that future research should critically reevaluate the current design choices of \tod systems to better tailor LLM-based \tod systems.

\rparagraph{Multilingual Generation with LLMs}
Our analysis demonstrates that when prompts are solely provided in English and LLMs are instructed to generate outputs in other languages, they often encounter difficulties in complying, resulting in the generation of outputs in English. Developing complex NLP applications like \tod systems requires a significant number of instructions to specify task requirements. However, the dominance of English instructions tends to bias model outputs towards English. Conversely, tailoring task instructions for each individual language, especially for resource-lean languages, poses a challenge. While some work has been recently conducted for other NLP tasks~\cite{bactrianx}, arguably less complex than \tod, the question of how to effectively control LLMs to generate target language outputs remains an open question for future research.

\section*{Limitations}

There are several limitations in this work, primarily stemming from the scope, design, and intended purpose of this toolkit. In this work, we only provide a set of baseline systems. As shown in Figure~\ref{fig:e2e_code}, these systems are intentionally kept as simple as possible while being fully functional, with the goal of enabling users to gain a conceptual understanding of the \tod task and implement their own systems with minimal learning effort. It is important to acknowledge that our systems may underperform compared to other more sophisticated systems, such as the one developed by~\citet{hudecek-dusek-2023-large} that employs advanced techniques for retrieving positive and negative ICL examples, or the system proposed by~\citet{zhang2023sgp} that incorporates an explicit and pre-defined task schema to guide system actions. Instead of pursuing state-of-the-art performance, we place greater emphasis on providing the essential environment and a set of tools, including automatic and human evaluation tools, to enable researchers to develop more advanced systems using our toolkit.

Currently, \toolkit currently supports only \tod datasets that are derived from the MultiWOZ dataset~\cite{Budzianowski:2018multiwoz} and its schema, that is, the ones annotated with the CUED schema~\cite{young2007cued}. We recognise the additional challenges associated with extending the toolkit to accommodate other datasets with different annotation schemata. Such extensions would typically involve the re-implementation of the data loader, the database, and some automatic evaluation metrics, such as Inform Rate and Success Rate. However, we believe that our human evaluation tool can be easily extended to evaluate systems developed on other datasets. 

Furthermore, it is worth noting that a fully inclusive dialogue system should consider not only text input but also other modalities, such as spoken and sign languages. We acknowledge that \toolkit currently focuses on text input only, and we hope to integrate the support for speech input and output as part of future work.

\section*{Ethics Statement}
The experimental study obtained the full Ethics Approval from the University of Cambridge in advance of its implementation and execution. Informed consent was obtained from every individual participant involved in the study, all of whom participated voluntarily. Our models leverage two data sources: the \dataset dataset and the pretraining data of each PLM employed in this study. \citet{multi3woz} highlights that the creation and publication of \dataset comply with the GDPR. Particularly, this dataset consists solely of hypothetical dialogues in which the domains and content have been restricted and predefined, minimising the risk of personal data being present. On the other hand, it is important to acknowledge that although these PLMs and LLMs are publicly available, there exists a potential risk of privacy violations~\cite{carlini2021extracting, 10.1145/3531146.3534642}.

\section*{Acknowledgements}
Songbo Hu is supported by the Cambridge International Scholarship.
Anna Korhonen acknowledges the support of the UK EPSRC grant EP/T02450X/1 and the UKRI Frontier grant EP/Y031350/1. Ivan Vuli\'{c} acknowledges the support of a personal Royal Society University Research Fellowship \textit{`Inclusive and Sustainable Language Technology for a Truly Multilingual World'} (no 221137; 2022--). We are grateful to our diligent participants for their efforts and contributions in our human evaluation experiment.

\bibliography{anthology,custom}

\begin{thebibliography}{52}
\expandafter\ifx\csname natexlab\endcsname\relax\def\natexlab#1{#1}\fi

\bibitem[{Banerjee and Lavie(2005)}]{banerjee2005meteor}
Satanjeev Banerjee and Alon Lavie. 2005.
\newblock \href {https://aclanthology.org/W05-0909} {{METEOR}: An automatic metric for {MT} evaluation with improved correlation with human judgments}.
\newblock In \emph{Proceedings of the {ACL} Workshop on Intrinsic and Extrinsic Evaluation Measures for Machine Translation and/or Summarization}, pages 65--72, Ann Arbor, Michigan. Association for Computational Linguistics.

\bibitem[{Bordes et~al.(2017)Bordes, Boureau, and Weston}]{bordes2017learning}
Antoine Bordes, Y-Lan Boureau, and Jason Weston. 2017.
\newblock \href {https://openreview.net/forum?id=S1Bb3D5gg} {Learning end-to-end goal-oriented dialog}.
\newblock In \emph{International Conference on Learning Representations}.

\bibitem[{Brown et~al.(2022)Brown, Lee, Mireshghallah, Shokri, and Tram\`{e}r}]{10.1145/3531146.3534642}
Hannah Brown, Katherine Lee, Fatemehsadat Mireshghallah, Reza Shokri, and Florian Tram\`{e}r. 2022.
\newblock \href {https://doi.org/10.1145/3531146.3534642} {What does it mean for a language model to preserve privacy?}
\newblock In \emph{Proceedings of the 2022 ACM Conference on Fairness, Accountability, and Transparency}, FAccT '22, page 2280–2292, New York, NY, USA. Association for Computing Machinery.

\bibitem[{Budzianowski et~al.(2018{\natexlab{a}})Budzianowski, Wen, Tseng, Casanueva, Ultes, Ramadan, and Ga{\v{s}}i{\'c}}]{budzianowski-etal-2018-multiwoz}
Pawe{\l} Budzianowski, Tsung-Hsien Wen, Bo-Hsiang Tseng, I{\~n}igo Casanueva, Stefan Ultes, Osman Ramadan, and Milica Ga{\v{s}}i{\'c}. 2018{\natexlab{a}}.
\newblock \href {https://doi.org/10.18653/v1/D18-1547} {{M}ulti{WOZ} - a large-scale multi-domain {W}izard-of-{O}z dataset for task-oriented dialogue modelling}.
\newblock In \emph{Proceedings of the 2018 Conference on Empirical Methods in Natural Language Processing}, pages 5016--5026, Brussels, Belgium. Association for Computational Linguistics.

\bibitem[{Budzianowski et~al.(2018{\natexlab{b}})Budzianowski, Wen, Tseng, Casanueva, Ultes, Ramadan, and Ga{\v{s}}i{\'c}}]{Budzianowski:2018multiwoz}
Pawe{\l} Budzianowski, Tsung-Hsien Wen, Bo-Hsiang Tseng, I{\~n}igo Casanueva, Stefan Ultes, Osman Ramadan, and Milica Ga{\v{s}}i{\'c}. 2018{\natexlab{b}}.
\newblock \href {https://doi.org/10.18653/v1/D18-1547} {{M}ulti{WOZ} - a large-scale multi-domain {W}izard-of-{O}z dataset for task-oriented dialogue modelling}.
\newblock In \emph{Proceedings of the 2018 Conference on Empirical Methods in Natural Language Processing}, pages 5016--5026, Brussels, Belgium. Association for Computational Linguistics.

\bibitem[{Byrne et~al.(2019)Byrne, Krishnamoorthi, Sankar, Neelakantan, Goodrich, Duckworth, Yavuz, Dubey, Kim, and Cedilnik}]{byrne-etal-2019-taskmaster}
Bill Byrne, Karthik Krishnamoorthi, Chinnadhurai Sankar, Arvind Neelakantan, Ben Goodrich, Daniel Duckworth, Semih Yavuz, Amit Dubey, Kyu-Young Kim, and Andy Cedilnik. 2019.
\newblock \href {https://doi.org/10.18653/v1/D19-1459} {Taskmaster-1: Toward a realistic and diverse dialog dataset}.
\newblock In \emph{Proceedings of the 2019 Conference on Empirical Methods in Natural Language Processing and the 9th International Joint Conference on Natural Language Processing (EMNLP-IJCNLP)}, pages 4516--4525, Hong Kong, China. Association for Computational Linguistics.

\bibitem[{Carlini et~al.(2021)Carlini, Tramer, Wallace, Jagielski, Herbert-Voss, Lee, Roberts, Brown, Song, Erlingsson et~al.}]{carlini2021extracting}
Nicholas Carlini, Florian Tramer, Eric Wallace, Matthew Jagielski, Ariel Herbert-Voss, Katherine Lee, Adam Roberts, Tom~B Brown, Dawn Song, Ulfar Erlingsson, et~al. 2021.
\newblock \href {https://www.usenix.org/system/files/sec21-carlini-extracting.pdf} {Extracting training data from large language models.}
\newblock In \emph{USENIX Security Symposium}, volume~6.

\bibitem[{Chung et~al.(2023)Chung, Cahyawijaya, Wilie, Lovenia, and Fung}]{chung2023instructtods}
Willy Chung, Samuel Cahyawijaya, Bryan Wilie, Holy Lovenia, and Pascale Fung. 2023.
\newblock Instructtods: Large language models for end-to-end task-oriented dialogue systems.
\newblock \emph{arXiv preprint arXiv:2310.08885}.

\bibitem[{Ding et~al.(2022)Ding, Hu, Bing, Aljunied, Joty, Si, and Miao}]{ding-etal-2022-globalwoz}
Bosheng Ding, Junjie Hu, Lidong Bing, Mahani Aljunied, Shafiq Joty, Luo Si, and Chunyan Miao. 2022.
\newblock \href {https://doi.org/10.18653/v1/2022.acl-long.115} {{G}lobal{W}o{Z}: Globalizing {M}ulti{W}o{Z} to develop multilingual task-oriented dialogue systems}.
\newblock In \emph{Proceedings of the 60th Annual Meeting of the Association for Computational Linguistics (Volume 1: Long Papers)}, pages 1639--1657, Dublin, Ireland. Association for Computational Linguistics.

\bibitem[{Eric et~al.(2017)Eric, Krishnan, Charette, and Manning}]{eric-etal-2017-key}
Mihail Eric, Lakshmi Krishnan, Francois Charette, and Christopher~D. Manning. 2017.
\newblock \href {https://doi.org/10.18653/v1/W17-5506} {Key-value retrieval networks for task-oriented dialogue}.
\newblock In \emph{Proceedings of the 18th Annual {SIG}dial Meeting on Discourse and Dialogue}, pages 37--49, Saarbr{\"u}cken, Germany. Association for Computational Linguistics.

\bibitem[{Eric and Manning(2017)}]{eric-manning-2017-copy}
Mihail Eric and Christopher Manning. 2017.
\newblock \href {https://aclanthology.org/E17-2075} {A copy-augmented sequence-to-sequence architecture gives good performance on task-oriented dialogue}.
\newblock In \emph{Proceedings of the 15th Conference of the {E}uropean Chapter of the Association for Computational Linguistics: Volume 2, Short Papers}, pages 468--473, Valencia, Spain. Association for Computational Linguistics.

\bibitem[{Gupta et~al.(2006)Gupta, T{\"{u}}r, Hakkani{-}T{\"{u}}r, Bangalore, Riccardi, and Gilbert}]{Gupta:2006}
Narendra~K. Gupta, G{\"{o}}khan T{\"{u}}r, Dilek Hakkani{-}T{\"{u}}r, Srinivas Bangalore, Giuseppe Riccardi, and Mazin Gilbert. 2006.
\newblock \href {https://doi.org/10.1109/TSA.2005.854085} {The at{\&}t spoken language understanding system}.
\newblock \emph{{IEEE} Transactions on Speech and Audio Processing}, 14(1):213--222.

\bibitem[{He et~al.(2022)He, Dai, Zheng, Wu, Cao, Liu, Jiang, Yang, Huang, Si et~al.}]{he2022galaxy}
Wanwei He, Yinpei Dai, Yinhe Zheng, Yuchuan Wu, Zheng Cao, Dermot Liu, Peng Jiang, Min Yang, Fei Huang, Luo Si, et~al. 2022.
\newblock \href {https://www.aaai.org/AAAI22Papers/AAAI-11845.HeW.pdf} {Galaxy: A generative pre-trained model for task-oriented dialog with semi-supervised learning and explicit policy injection}.
\newblock \emph{Proceedings of the AAAI Conference on Artificial Intelligence}.

\bibitem[{Heck et~al.(2023)Heck, Lubis, Ruppik, Vukovic, Feng, Geishauser, Lin, van Niekerk, and Gasic}]{heck2023chatgpt}
Michael Heck, Nurul Lubis, Benjamin Ruppik, Renato Vukovic, Shutong Feng, Christian Geishauser, Hsien-chin Lin, Carel van Niekerk, and Milica Gasic. 2023.
\newblock \href {https://doi.org/10.18653/v1/2023.acl-short.81} {{C}hat{GPT} for zero-shot dialogue state tracking: A solution or an opportunity?}
\newblock In \emph{Proceedings of the 61st Annual Meeting of the Association for Computational Linguistics (Volume 2: Short Papers)}, pages 936--950, Toronto, Canada. Association for Computational Linguistics.

\bibitem[{Hu et~al.(2023{\natexlab{a}})Hu, Zhou, Hergul, Gritta, Zhang, Iacobacci, Vuli{\'c}, and Korhonen}]{multi3woz}
Songbo Hu, Han Zhou, Mete Hergul, Milan Gritta, Guchun Zhang, Ignacio Iacobacci, Ivan Vuli{\'c}, and Anna Korhonen. 2023{\natexlab{a}}.
\newblock \href {https://doi.org/10.1162/tacl_a_00609} {Multi 3 {WOZ}: A multilingual, multi-domain, multi-parallel dataset for training and evaluating culturally adapted task-oriented dialog systems}.
\newblock \emph{Transactions of the Association for Computational Linguistics}, 11:1396--1415.

\bibitem[{Hu et~al.(2023{\natexlab{b}})Hu, Zhou, Yuan, Gritta, Zhang, Iacobacci, Korhonen, and Vuli{\'c}}]{hu2023systematic}
Songbo Hu, Han Zhou, Moy Yuan, Milan Gritta, Guchun Zhang, Ignacio Iacobacci, Anna Korhonen, and Ivan Vuli{\'c}. 2023{\natexlab{b}}.
\newblock \href {https://aclanthology.org/2023.emnlp-main.422} {A systematic study of performance disparities in multilingual task-oriented dialogue systems}.
\newblock In \emph{Proceedings of the 2023 Conference on Empirical Methods in Natural Language Processing}, pages 6825--6851, Singapore. Association for Computational Linguistics.

\bibitem[{Hude{\v{c}}ek and Dusek(2023)}]{hudecek-dusek-2023-large}
Vojt{\v{e}}ch Hude{\v{c}}ek and Ondrej Dusek. 2023.
\newblock \href {https://aclanthology.org/2023.sigdial-1.21} {Are large language models all you need for task-oriented dialogue?}
\newblock In \emph{Proceedings of the 24th Meeting of the Special Interest Group on Discourse and Dialogue}, pages 216--228, Prague, Czechia. Association for Computational Linguistics.

\bibitem[{Huynh et~al.(2022)Huynh, Chiang, Bigham, and Eskenazi}]{huynh-etal-2022-dialcrowd}
Jessica Huynh, Ting-Rui Chiang, Jeffrey Bigham, and Maxine Eskenazi. 2022.
\newblock \href {https://aclanthology.org/2022.lrec-1.134} {{D}ial{C}rowd 2.0: A quality-focused dialog system crowdsourcing toolkit}.
\newblock In \emph{Proceedings of the Thirteenth Language Resources and Evaluation Conference}, pages 1256--1263, Marseille, France. European Language Resources Association.

\bibitem[{Lee et~al.(2018)Lee, Zhao, Black, and Eskenazi}]{lee-etal-2018-dialcrowd}
Kyusong Lee, Tiancheng Zhao, Alan~W. Black, and Maxine Eskenazi. 2018.
\newblock \href {https://doi.org/10.18653/v1/W18-5028} {{D}ial{C}rowd: A toolkit for easy dialog system assessment}.
\newblock In \emph{Proceedings of the 19th Annual {SIG}dial Meeting on Discourse and Dialogue}, pages 245--248, Melbourne, Australia. Association for Computational Linguistics.

\bibitem[{Lee et~al.(2019)Lee, Zhu, Takanobu, Zhang, Zhang, Li, Li, Peng, Li, Huang, and Gao}]{lee-etal-2019-convlab}
Sungjin Lee, Qi~Zhu, Ryuichi Takanobu, Zheng Zhang, Yaoqin Zhang, Xiang Li, Jinchao Li, Baolin Peng, Xiujun Li, Minlie Huang, and Jianfeng Gao. 2019.
\newblock \href {https://doi.org/10.18653/v1/P19-3011} {{C}onv{L}ab: Multi-domain end-to-end dialog system platform}.
\newblock In \emph{Proceedings of the 57th Annual Meeting of the Association for Computational Linguistics: System Demonstrations}, pages 64--69, Florence, Italy. Association for Computational Linguistics.

\bibitem[{Lei et~al.(2018)Lei, Jin, Kan, Ren, He, and Yin}]{lei2018sequicity}
Wenqiang Lei, Xisen Jin, Min-Yen Kan, Zhaochun Ren, Xiangnan He, and Dawei Yin. 2018.
\newblock \href {https://doi.org/10.18653/v1/P18-1133} {{S}equicity: Simplifying task-oriented dialogue systems with single sequence-to-sequence architectures}.
\newblock In \emph{Proceedings of the 56th Annual Meeting of the Association for Computational Linguistics (Volume 1: Long Papers)}, pages 1437--1447, Melbourne, Australia. Association for Computational Linguistics.

\bibitem[{Lewis et~al.(2020)Lewis, Liu, Goyal, Ghazvininejad, Mohamed, Levy, Stoyanov, and Zettlemoyer}]{lewis-etal-2020-bart}
Mike Lewis, Yinhan Liu, Naman Goyal, Marjan Ghazvininejad, Abdelrahman Mohamed, Omer Levy, Veselin Stoyanov, and Luke Zettlemoyer. 2020.
\newblock \href {https://doi.org/10.18653/v1/2020.acl-main.703} {{BART}: Denoising sequence-to-sequence pre-training for natural language generation, translation, and comprehension}.
\newblock In \emph{Proceedings of the 58th Annual Meeting of the Association for Computational Linguistics}, pages 7871--7880, Online. Association for Computational Linguistics.

\bibitem[{Li et~al.(2023)Li, Koto, Wu, Aji, and Baldwin}]{bactrianx}
Haonan Li, Fajri Koto, Minghao Wu, Alham~Fikri Aji, and Timothy Baldwin. 2023.
\newblock \href {https://doi.org/10.48550/arXiv.2305.15011} {{Bactrian-X:} {A} multilingual replicable instruction-following model with low-rank adaptation}.
\newblock \emph{CoRR}, abs/2305.15011.

\bibitem[{Lin(2004)}]{lin-2004-rouge}
Chin-Yew Lin. 2004.
\newblock \href {https://aclanthology.org/W04-1013} {{ROUGE}: A package for automatic evaluation of summaries}.
\newblock In \emph{Text Summarization Branches Out}, pages 74--81, Barcelona, Spain. Association for Computational Linguistics.

\bibitem[{Lin et~al.(2020)Lin, Madotto, Winata, and Fung}]{lin2020mintl}
Zhaojiang Lin, Andrea Madotto, Genta~Indra Winata, and Pascale Fung. 2020.
\newblock \href {https://doi.org/10.18653/v1/2020.emnlp-main.273} {{M}in{TL}: Minimalist transfer learning for task-oriented dialogue systems}.
\newblock In \emph{Proceedings of the 2020 Conference on Empirical Methods in Natural Language Processing (EMNLP)}, pages 3391--3405, Online. Association for Computational Linguistics.

\bibitem[{Mehri et~al.(2022)Mehri, Choi, D'Haro, Deriu, Eskenazi, Gasic, Georgila, Hakkani-Tur, Li, Rieser, Shaikh, Traum, Yeh, Yu, Zhang, and Zhang}]{mehri2022report}
Shikib Mehri, Jinho Choi, Luis~Fernando D'Haro, Jan Deriu, Maxine Eskenazi, Milica Gasic, Kallirroi Georgila, Dilek Hakkani-Tur, Zekang Li, Verena Rieser, Samira Shaikh, David Traum, Yi-Ting Yeh, Zhou Yu, Yizhe Zhang, and Chen Zhang. 2022.
\newblock \href {http://arxiv.org/abs/2203.10012} {Report from the nsf future directions workshop on automatic evaluation of dialog: Research directions and challenges}.

\bibitem[{Mehri and Eskenazi(2020)}]{mehri-eskenazi-2020-unsupervised}
Shikib Mehri and Maxine Eskenazi. 2020.
\newblock \href {https://aclanthology.org/2020.sigdial-1.28} {Unsupervised evaluation of interactive dialog with {D}ialo{GPT}}.
\newblock In \emph{Proceedings of the 21th Annual Meeting of the Special Interest Group on Discourse and Dialogue}, pages 225--235, 1st virtual meeting. Association for Computational Linguistics.

\bibitem[{Nakatani(2010)}]{nakatani2010langdetect}
Shuyo Nakatani. 2010.
\newblock \href {https://github.com/shuyo/language-detection} {Language detection library for java}.

\bibitem[{Nekvinda and Du{\v{s}}ek(2021)}]{nekvinda-dusek-2021-shades}
Tom{\'a}{\v{s}} Nekvinda and Ond{\v{r}}ej Du{\v{s}}ek. 2021.
\newblock \href {https://doi.org/10.18653/v1/2021.gem-1.4} {Shades of {BLEU}, flavours of success: The case of {M}ulti{WOZ}}.
\newblock In \emph{Proceedings of the 1st Workshop on Natural Language Generation, Evaluation, and Metrics (GEM 2021)}, pages 34--46, Online. Association for Computational Linguistics.

\bibitem[{OpenAI(2023)}]{OpenAI2023GPT4TR}
OpenAI. 2023.
\newblock \href {https://api.semanticscholar.org/CorpusID:257532815} {Gpt-4 technical report}.
\newblock \emph{ArXiv}, abs/2303.08774.

\bibitem[{Ouyang et~al.(2022)Ouyang, Wu, Jiang, Almeida, Wainwright, Mishkin, Zhang, Agarwal, Slama, Ray et~al.}]{ouyang2022training}
Long Ouyang, Jeffrey Wu, Xu~Jiang, Diogo Almeida, Carroll Wainwright, Pamela Mishkin, Chong Zhang, Sandhini Agarwal, Katarina Slama, Alex Ray, et~al. 2022.
\newblock Training language models to follow instructions with human feedback.
\newblock \emph{Advances in Neural Information Processing Systems}, 35:27730--27744.

\bibitem[{Papineni et~al.(2002)Papineni, Roukos, Ward, and Zhu}]{papineni2002bleu}
Kishore Papineni, Salim Roukos, Todd Ward, and Wei-Jing Zhu. 2002.
\newblock \href {https://doi.org/10.3115/1073083.1073135} {{B}leu: a method for automatic evaluation of machine translation}.
\newblock In \emph{Proceedings of the 40th Annual Meeting of the Association for Computational Linguistics}, pages 311--318, Philadelphia, Pennsylvania, USA. Association for Computational Linguistics.

\bibitem[{Peng et~al.(2021)Peng, Li, Li, Shayandeh, Liden, and Gao}]{peng-etal-2021-soloist}
Baolin Peng, Chunyuan Li, Jinchao Li, Shahin Shayandeh, Lars Liden, and Jianfeng Gao. 2021.
\newblock \href {https://doi.org/10.1162/tacl_a_00399} {Soloist: Building task bots at scale with transfer learning and machine teaching}.
\newblock \emph{Transactions of the Association for Computational Linguistics}, 9:807--824.

\bibitem[{Raffel et~al.(2020)Raffel, Shazeer, Roberts, Lee, Narang, Matena, Zhou, Li, and Liu}]{t52020}
Colin Raffel, Noam Shazeer, Adam Roberts, Katherine Lee, Sharan Narang, Michael Matena, Yanqi Zhou, Wei Li, and Peter~J. Liu. 2020.
\newblock Exploring the limits of transfer learning with a unified text-to-text transformer.
\newblock \emph{J. Mach. Learn. Res.}, 21(1).

\bibitem[{Razumovskaia et~al.(2022)Razumovskaia, Glava\v{s}, Majewska, Ponti, Korhonen, and Vuli\'{c}}]{Razumovskaia:2022survey}
Evgeniia Razumovskaia, Goran Glava\v{s}, Olga Majewska, Edoardo~Maria Ponti, Anna Korhonen, and Ivan Vuli\'{c}. 2022.
\newblock \href {https://doi.org/10.1613/jair.1.13083} {Crossing the conversational chasm: {A} primer on natural language processing for multilingual task-oriented dialogue systems}.
\newblock \emph{Journal of Artificial Intelligence Research}, 74:1351--1402.

\bibitem[{Srivastava et~al.(2023)Srivastava, Rastogi, Rao, Shoeb, Abid, Fisch, Brown, Santoro, Gupta, Garriga-Alonso, Kluska, Lewkowycz, Agarwal, Power, Ray, Warstadt, Kocurek, Safaya, Tazarv, Xiang, Parrish, Nie, Hussain, Askell, Dsouza, Slone, Rahane, Iyer, Andreassen, Madotto, Santilli, Stuhlm{\"u}ller, Dai, La, Lampinen, Zou, Jiang, Chen, Vuong, Gupta, Gottardi, Norelli, Venkatesh, Gholamidavoodi, Tabassum, Menezes, Kirubarajan, Mullokandov, Sabharwal, Herrick, Efrat, Erdem, Karaka{\c{s}}, Roberts, Loe, Zoph, Bojanowski, {\"O}zyurt, Hedayatnia, Neyshabur, Inden, Stein, Ekmekci, Lin, Howald, Orinion, Diao, Dour, Stinson, Argueta, Ferri, Singh, Rathkopf, Meng, Baral, Wu, Callison-Burch, Waites, Voigt, Manning, Potts, Ramirez, Rivera, Siro, Raffel, Ashcraft, Garbacea, Sileo, Garrette, Hendrycks, Kilman, Roth, Freeman, Khashabi, Levy, Gonz{\'a}lez, Perszyk, Hernandez, Chen, Ippolito, Gilboa, Dohan, Drakard, Jurgens, Datta, Ganguli, Emelin, Kleyko, Yuret, Chen, Tam, Hupkes, Misra, Buzan, Mollo, Yang, Lee,
  Schrader, Shutova, Cubuk, Segal, Hagerman, Barnes, Donoway, Pavlick, Rodol{\`a}, Lam, Chu, Tang, Erdem, Chang, Chi, Dyer, Jerzak, Kim, Manyasi, Zheltonozhskii, Xia, Siar, Mart{\'\i}nez-Plumed, Happ{\'e}, Chollet, Rong, Mishra, Winata, de~Melo, Kruszewski, Parascandolo, Mariani, Wang, Jaimovitch-Lopez, Betz, Gur-Ari, Galijasevic, Kim, Rashkin, Hajishirzi, Mehta, Bogar, Shevlin, Schuetze, Yakura, Zhang, Wong, Ng, Noble, Jumelet, Geissinger, Kernion, Hilton, Lee, Fisac, Simon, Koppel, Zheng, Zou, Kocon, Thompson, Wingfield, Kaplan, Radom, Sohl-Dickstein, Phang, Wei, Yosinski, Novikova, Bosscher, Marsh, Kim, Taal, Engel, Alabi, Xu, Song, Tang, Waweru, Burden, Miller, Balis, Batchelder, Berant, Frohberg, Rozen, Hernandez-Orallo, Boudeman, Guerr, Jones, Tenenbaum, Rule, Chua, Kanclerz, Livescu, Krauth, Gopalakrishnan, Ignatyeva, Markert, Dhole, Gimpel, Omondi, Mathewson, Chiafullo, Shkaruta, Shridhar, McDonell, Richardson, Reynolds, Gao, Zhang, Dugan, Qin, Contreras-Ochando, Morency, Moschella, Lam, Noble,
  Schmidt, He, Oliveros-Col{\'o}n, Metz, Senel, Bosma, Sap, Hoeve, Farooqi, Faruqui, Mazeika, Baturan, Marelli, Maru, Ramirez-Quintana, Tolkiehn, Giulianelli, Lewis, Potthast, Leavitt, Hagen, Schubert, Baitemirova, Arnaud, McElrath, Yee, Cohen, Gu, Ivanitskiy, Starritt, Strube, Sw{\k{e}}drowski, Bevilacqua, Yasunaga, Kale, Cain, Xu, Suzgun, Walker, Tiwari, Bansal, Aminnaseri, Geva, Gheini, T, Peng, Chi, Lee, Krakover, Cameron, Roberts, Doiron, Martinez, Nangia, Deckers, Muennighoff, Keskar, Iyer, Constant, Fiedel, Wen, Zhang, Agha, Elbaghdadi, Levy, Evans, Casares, Doshi, Fung, Liang, Vicol, Alipoormolabashi, Liao, Liang, Chang, Eckersley, Htut, Hwang, Mi{\l}kowski, Patil, Pezeshkpour, Oli, Mei, Lyu, Chen, Banjade, Rudolph, Gabriel, Habacker, Risco, Milli{\`e}re, Garg, Barnes, Saurous, Arakawa, Raymaekers, Frank, Sikand, Novak, Sitelew, Bras, Liu, Jacobs, Zhang, Salakhutdinov, Chi, Lee, Stovall, Teehan, Yang, Singh, Mohammad, Anand, Dillavou, Shleifer, Wiseman, Gruetter, Bowman, Schoenholz, Han, Kwatra, Rous,
  Ghazarian, Ghosh, Casey, Bischoff, Gehrmann, Schuster, Sadeghi, Hamdan, Zhou, Srivastava, Shi, Singh, Asaadi, Gu, Pachchigar, Toshniwal, Upadhyay, Debnath, Shakeri, Thormeyer, Melzi, Reddy, Makini, Lee, Torene, Hatwar, Dehaene, Divic, Ermon, Biderman, Lin, Prasad, Piantadosi, Shieber, Misherghi, Kiritchenko, Mishra, Linzen, Schuster, Li, Yu, Ali, Hashimoto, Wu, Desbordes, Rothschild, Phan, Wang, Nkinyili, Schick, Kornev, Tunduny, Gerstenberg, Chang, Neeraj, Khot, Shultz, Shaham, Misra, Demberg, Nyamai, Raunak, Ramasesh, vinay~uday prabhu, Padmakumar, Srikumar, Fedus, Saunders, Zhang, Vossen, Ren, Tong, Zhao, Wu, Shen, Yaghoobzadeh, Lakretz, Song, Bahri, Choi, Yang, Hao, Chen, Belinkov, Hou, Hou, Bai, Seid, Zhao, Wang, Wang, Wang, and Wu}]{srivastava2023beyond}
Aarohi Srivastava, Abhinav Rastogi, Abhishek Rao, Abu Awal~Md Shoeb, Abubakar Abid, Adam Fisch, Adam~R. Brown, Adam Santoro, Aditya Gupta, Adri{\`a} Garriga-Alonso, Agnieszka Kluska, Aitor Lewkowycz, Akshat Agarwal, Alethea Power, Alex Ray, Alex Warstadt, Alexander~W. Kocurek, Ali Safaya, Ali Tazarv, Alice Xiang, Alicia Parrish, Allen Nie, Aman Hussain, Amanda Askell, Amanda Dsouza, Ambrose Slone, Ameet Rahane, Anantharaman~S. Iyer, Anders~Johan Andreassen, Andrea Madotto, Andrea Santilli, Andreas Stuhlm{\"u}ller, Andrew~M. Dai, Andrew La, Andrew Lampinen, Andy Zou, Angela Jiang, Angelica Chen, Anh Vuong, Animesh Gupta, Anna Gottardi, Antonio Norelli, Anu Venkatesh, Arash Gholamidavoodi, Arfa Tabassum, Arul Menezes, Arun Kirubarajan, Asher Mullokandov, Ashish Sabharwal, Austin Herrick, Avia Efrat, Aykut Erdem, Ayla Karaka{\c{s}}, B.~Ryan Roberts, Bao~Sheng Loe, Barret Zoph, Bart{\l}omiej Bojanowski, Batuhan {\"O}zyurt, Behnam Hedayatnia, Behnam Neyshabur, Benjamin Inden, Benno Stein, Berk Ekmekci, Bill~Yuchen
  Lin, Blake Howald, Bryan Orinion, Cameron Diao, Cameron Dour, Catherine Stinson, Cedrick Argueta, Cesar Ferri, Chandan Singh, Charles Rathkopf, Chenlin Meng, Chitta Baral, Chiyu Wu, Chris Callison-Burch, Christopher Waites, Christian Voigt, Christopher~D Manning, Christopher Potts, Cindy Ramirez, Clara~E. Rivera, Clemencia Siro, Colin Raffel, Courtney Ashcraft, Cristina Garbacea, Damien Sileo, Dan Garrette, Dan Hendrycks, Dan Kilman, Dan Roth, C.~Daniel Freeman, Daniel Khashabi, Daniel Levy, Daniel~Mosegu{\'\i} Gonz{\'a}lez, Danielle Perszyk, Danny Hernandez, Danqi Chen, Daphne Ippolito, Dar Gilboa, David Dohan, David Drakard, David Jurgens, Debajyoti Datta, Deep Ganguli, Denis Emelin, Denis Kleyko, Deniz Yuret, Derek Chen, Derek Tam, Dieuwke Hupkes, Diganta Misra, Dilyar Buzan, Dimitri~Coelho Mollo, Diyi Yang, Dong-Ho Lee, Dylan Schrader, Ekaterina Shutova, Ekin~Dogus Cubuk, Elad Segal, Eleanor Hagerman, Elizabeth Barnes, Elizabeth Donoway, Ellie Pavlick, Emanuele Rodol{\`a}, Emma Lam, Eric Chu, Eric Tang,
  Erkut Erdem, Ernie Chang, Ethan~A Chi, Ethan Dyer, Ethan Jerzak, Ethan Kim, Eunice~Engefu Manyasi, Evgenii Zheltonozhskii, Fanyue Xia, Fatemeh Siar, Fernando Mart{\'\i}nez-Plumed, Francesca Happ{\'e}, Francois Chollet, Frieda Rong, Gaurav Mishra, Genta~Indra Winata, Gerard de~Melo, Germ{\'a}n Kruszewski, Giambattista Parascandolo, Giorgio Mariani, Gloria~Xinyue Wang, Gonzalo Jaimovitch-Lopez, Gregor Betz, Guy Gur-Ari, Hana Galijasevic, Hannah Kim, Hannah Rashkin, Hannaneh Hajishirzi, Harsh Mehta, Hayden Bogar, Henry Francis~Anthony Shevlin, Hinrich Schuetze, Hiromu Yakura, Hongming Zhang, Hugh~Mee Wong, Ian Ng, Isaac Noble, Jaap Jumelet, Jack Geissinger, Jackson Kernion, Jacob Hilton, Jaehoon Lee, Jaime~Fern{\'a}ndez Fisac, James~B Simon, James Koppel, James Zheng, James Zou, Jan Kocon, Jana Thompson, Janelle Wingfield, Jared Kaplan, Jarema Radom, Jascha Sohl-Dickstein, Jason Phang, Jason Wei, Jason Yosinski, Jekaterina Novikova, Jelle Bosscher, Jennifer Marsh, Jeremy Kim, Jeroen Taal, Jesse Engel, Jesujoba
  Alabi, Jiacheng Xu, Jiaming Song, Jillian Tang, Joan Waweru, John Burden, John Miller, John~U. Balis, Jonathan Batchelder, Jonathan Berant, J{\"o}rg Frohberg, Jos Rozen, Jose Hernandez-Orallo, Joseph Boudeman, Joseph Guerr, Joseph Jones, Joshua~B. Tenenbaum, Joshua~S. Rule, Joyce Chua, Kamil Kanclerz, Karen Livescu, Karl Krauth, Karthik Gopalakrishnan, Katerina Ignatyeva, Katja Markert, Kaustubh Dhole, Kevin Gimpel, Kevin Omondi, Kory~Wallace Mathewson, Kristen Chiafullo, Ksenia Shkaruta, Kumar Shridhar, Kyle McDonell, Kyle Richardson, Laria Reynolds, Leo Gao, Li~Zhang, Liam Dugan, Lianhui Qin, Lidia Contreras-Ochando, Louis-Philippe Morency, Luca Moschella, Lucas Lam, Lucy Noble, Ludwig Schmidt, Luheng He, Luis Oliveros-Col{\'o}n, Luke Metz, L{\"u}tfi~Kerem Senel, Maarten Bosma, Maarten Sap, Maartje~Ter Hoeve, Maheen Farooqi, Manaal Faruqui, Mantas Mazeika, Marco Baturan, Marco Marelli, Marco Maru, Maria~Jose Ramirez-Quintana, Marie Tolkiehn, Mario Giulianelli, Martha Lewis, Martin Potthast, Matthew~L
  Leavitt, Matthias Hagen, M{\'a}ty{\'a}s Schubert, Medina~Orduna Baitemirova, Melody Arnaud, Melvin McElrath, Michael~Andrew Yee, Michael Cohen, Michael Gu, Michael Ivanitskiy, Michael Starritt, Michael Strube, Micha{\l} Sw{\k{e}}drowski, Michele Bevilacqua, Michihiro Yasunaga, Mihir Kale, Mike Cain, Mimee Xu, Mirac Suzgun, Mitch Walker, Mo~Tiwari, Mohit Bansal, Moin Aminnaseri, Mor Geva, Mozhdeh Gheini, Mukund~Varma T, Nanyun Peng, Nathan~Andrew Chi, Nayeon Lee, Neta Gur-Ari Krakover, Nicholas Cameron, Nicholas Roberts, Nick Doiron, Nicole Martinez, Nikita Nangia, Niklas Deckers, Niklas Muennighoff, Nitish~Shirish Keskar, Niveditha~S. Iyer, Noah Constant, Noah Fiedel, Nuan Wen, Oliver Zhang, Omar Agha, Omar Elbaghdadi, Omer Levy, Owain Evans, Pablo Antonio~Moreno Casares, Parth Doshi, Pascale Fung, Paul~Pu Liang, Paul Vicol, Pegah Alipoormolabashi, Peiyuan Liao, Percy Liang, Peter~W Chang, Peter Eckersley, Phu~Mon Htut, Pinyu Hwang, Piotr Mi{\l}kowski, Piyush Patil, Pouya Pezeshkpour, Priti Oli, Qiaozhu
  Mei, Qing Lyu, Qinlang Chen, Rabin Banjade, Rachel~Etta Rudolph, Raefer Gabriel, Rahel Habacker, Ramon Risco, Rapha{\"e}l Milli{\`e}re, Rhythm Garg, Richard Barnes, Rif~A. Saurous, Riku Arakawa, Robbe Raymaekers, Robert Frank, Rohan Sikand, Roman Novak, Roman Sitelew, Ronan~Le Bras, Rosanne Liu, Rowan Jacobs, Rui Zhang, Russ Salakhutdinov, Ryan~Andrew Chi, Seungjae~Ryan Lee, Ryan Stovall, Ryan Teehan, Rylan Yang, Sahib Singh, Saif~M. Mohammad, Sajant Anand, Sam Dillavou, Sam Shleifer, Sam Wiseman, Samuel Gruetter, Samuel~R. Bowman, Samuel~Stern Schoenholz, Sanghyun Han, Sanjeev Kwatra, Sarah~A. Rous, Sarik Ghazarian, Sayan Ghosh, Sean Casey, Sebastian Bischoff, Sebastian Gehrmann, Sebastian Schuster, Sepideh Sadeghi, Shadi Hamdan, Sharon Zhou, Shashank Srivastava, Sherry Shi, Shikhar Singh, Shima Asaadi, Shixiang~Shane Gu, Shubh Pachchigar, Shubham Toshniwal, Shyam Upadhyay, Shyamolima~Shammie Debnath, Siamak Shakeri, Simon Thormeyer, Simone Melzi, Siva Reddy, Sneha~Priscilla Makini, Soo-Hwan Lee, Spencer
  Torene, Sriharsha Hatwar, Stanislas Dehaene, Stefan Divic, Stefano Ermon, Stella Biderman, Stephanie Lin, Stephen Prasad, Steven Piantadosi, Stuart Shieber, Summer Misherghi, Svetlana Kiritchenko, Swaroop Mishra, Tal Linzen, Tal Schuster, Tao Li, Tao Yu, Tariq Ali, Tatsunori Hashimoto, Te-Lin Wu, Th{\'e}o Desbordes, Theodore Rothschild, Thomas Phan, Tianle Wang, Tiberius Nkinyili, Timo Schick, Timofei Kornev, Titus Tunduny, Tobias Gerstenberg, Trenton Chang, Trishala Neeraj, Tushar Khot, Tyler Shultz, Uri Shaham, Vedant Misra, Vera Demberg, Victoria Nyamai, Vikas Raunak, Vinay~Venkatesh Ramasesh, vinay~uday prabhu, Vishakh Padmakumar, Vivek Srikumar, William Fedus, William Saunders, William Zhang, Wout Vossen, Xiang Ren, Xiaoyu Tong, Xinran Zhao, Xinyi Wu, Xudong Shen, Yadollah Yaghoobzadeh, Yair Lakretz, Yangqiu Song, Yasaman Bahri, Yejin Choi, Yichi Yang, Yiding Hao, Yifu Chen, Yonatan Belinkov, Yu~Hou, Yufang Hou, Yuntao Bai, Zachary Seid, Zhuoye Zhao, Zijian Wang, Zijie~J. Wang, Zirui Wang, and Ziyi Wu.
  2023.
\newblock \href {https://openreview.net/forum?id=uyTL5Bvosj} {Beyond the imitation game: Quantifying and extrapolating the capabilities of language models}.
\newblock \emph{Transactions on Machine Learning Research}.

\bibitem[{Touvron et~al.(2023)Touvron, Lavril, Izacard, Martinet, Lachaux, Lacroix, Rozi{\`e}re, Goyal, Hambro, Azhar et~al.}]{touvron2023llama}
Hugo Touvron, Thibaut Lavril, Gautier Izacard, Xavier Martinet, Marie-Anne Lachaux, Timoth{\'e}e Lacroix, Baptiste Rozi{\`e}re, Naman Goyal, Eric Hambro, Faisal Azhar, et~al. 2023.
\newblock Llama: Open and efficient foundation language models.
\newblock \emph{arXiv preprint arXiv:2302.13971}.

\bibitem[{T{\"{u}}r et~al.(2010)T{\"{u}}r, Stolcke, Voss, Peters, Hakkani{-}T{\"{u}}r, Dowding, Favre, Fern{\'{a}}ndez, Frampton, Frandsen, Frederickson, Graciarena, Kintzing, Leveque, Mason, Niekrasz, Purver, Riedhammer, Shriberg, Tien, Vergyri, and Yang}]{Tur:2010}
G{\"{o}}khan T{\"{u}}r, Andreas Stolcke, L.~Lynn Voss, Stanley Peters, Dilek Hakkani{-}T{\"{u}}r, John Dowding, Beno{\^{\i}}t Favre, Raquel Fern{\'{a}}ndez, Matthew Frampton, Michael~W. Frandsen, Clint Frederickson, Martin Graciarena, Donald Kintzing, Kyle Leveque, Shane Mason, John Niekrasz, Matthew Purver, Korbinian Riedhammer, Elizabeth Shriberg, Jing Tien, Dimitra Vergyri, and Fan Yang. 2010.
\newblock \href {https://doi.org/10.1109/TASL.2009.2038810} {The {CALO} meeting assistant system}.
\newblock \emph{{IEEE} Transactions on Speech Audio Processing}, 18(6):1601--1611.

\bibitem[{Ultes et~al.(2017)Ultes, Rojas~Barahona, Su, Vandyke, Kim, Casanueva, Budzianowski, Mrk\v{s}i\'{c}, Wen, Gasic, and Young}]{ultes2017pydial}
Stefan Ultes, Lina~M. Rojas~Barahona, Pei-Hao Su, David Vandyke, Dongho Kim, I\~{n}igo Casanueva, Pawe{\l} Budzianowski, Nikola Mrk\v{s}i\'{c}, Tsung-Hsien Wen, Milica Gasic, and Steve Young. 2017.
\newblock \href {http://aclweb.org/anthology/P17-4013} {{PyDial: A Multi-domain Statistical Dialogue System Toolkit}}.
\newblock In \emph{Proceedings of ACL 2017, System Demonstrations}, pages 73--78, Vancouver, Canada. Association for Computational Linguistics.

\bibitem[{Wang et~al.(2019)Wang, Pruksachatkun, Nangia, Singh, Michael, Hill, Levy, and Bowman}]{superglue}
Alex Wang, Yada Pruksachatkun, Nikita Nangia, Amanpreet Singh, Julian Michael, Felix Hill, Omer Levy, and Samuel~R. Bowman. 2019.
\newblock \emph{SuperGLUE: A Stickier Benchmark for General-Purpose Language Understanding Systems}. Curran Associates Inc., Red Hook, NY, USA.

\bibitem[{Wang et~al.(2023)Wang, Cheng, Zhan, Li, Song, and Liu}]{wang2023openchat}
Guan Wang, Sijie Cheng, Xianyuan Zhan, Xiangang Li, Sen Song, and Yang Liu. 2023.
\newblock Openchat: Advancing open-source language models with mixed-quality data.
\newblock \emph{arXiv preprint arXiv:2309.11235}.

\bibitem[{Wang et~al.(2022)Wang, Mishra, Alipoormolabashi, Kordi, Mirzaei, Arunkumar, Ashok, Dhanasekaran, Naik, Stap et~al.}]{wang2022super}
Yizhong Wang, Swaroop Mishra, Pegah Alipoormolabashi, Yeganeh Kordi, Amirreza Mirzaei, Anjana Arunkumar, Arjun Ashok, Arut~Selvan Dhanasekaran, Atharva Naik, David Stap, et~al. 2022.
\newblock Super-naturalinstructions: Generalization via declarative instructions on 1600+ nlp tasks.
\newblock \emph{arXiv preprint arXiv:2204.07705}.

\bibitem[{Wen et~al.(2017)Wen, Vandyke, Mrk{\v{s}}i{\'c}, Ga{\v{s}}i{\'c}, Rojas-Barahona, Su, Ultes, and Young}]{wen2016network}
Tsung-Hsien Wen, David Vandyke, Nikola Mrk{\v{s}}i{\'c}, Milica Ga{\v{s}}i{\'c}, Lina~M. Rojas-Barahona, Pei-Hao Su, Stefan Ultes, and Steve Young. 2017.
\newblock \href {https://aclanthology.org/E17-1042} {A network-based end-to-end trainable task-oriented dialogue system}.
\newblock In \emph{Proceedings of the 15th Conference of the {E}uropean Chapter of the Association for Computational Linguistics: Volume 1, Long Papers}, pages 438--449, Valencia, Spain. Association for Computational Linguistics.

\bibitem[{Wolf et~al.(2020)Wolf, Debut, Sanh, Chaumond, Delangue, Moi, Cistac, Rault, Louf, Funtowicz, Davison, Shleifer, von Platen, Ma, Jernite, Plu, Xu, Le~Scao, Gugger, Drame, Lhoest, and Rush}]{wolf2019huggingface}
Thomas Wolf, Lysandre Debut, Victor Sanh, Julien Chaumond, Clement Delangue, Anthony Moi, Pierric Cistac, Tim Rault, Remi Louf, Morgan Funtowicz, Joe Davison, Sam Shleifer, Patrick von Platen, Clara Ma, Yacine Jernite, Julien Plu, Canwen Xu, Teven Le~Scao, Sylvain Gugger, Mariama Drame, Quentin Lhoest, and Alexander Rush. 2020.
\newblock \href {https://doi.org/10.18653/v1/2020.emnlp-demos.6} {Transformers: State-of-the-art natural language processing}.
\newblock In \emph{Proceedings of the 2020 Conference on Empirical Methods in Natural Language Processing: System Demonstrations}, pages 38--45, Online. Association for Computational Linguistics.

\bibitem[{Wu et~al.(2023)Wu, Alnuhait, Chen, and Yu}]{wu2023using}
Qingyang Wu, Deema Alnuhait, Derek Chen, and Zhou Yu. 2023.
\newblock Using textual interface to align external knowledge for end-to-end task-oriented dialogue systems.
\newblock \emph{arXiv preprint arXiv:2305.13710}.

\bibitem[{Xue et~al.(2021)Xue, Constant, Roberts, Kale, Al-Rfou, Siddhant, Barua, and Raffel}]{xue-etal-2021-mt5}
Linting Xue, Noah Constant, Adam Roberts, Mihir Kale, Rami Al-Rfou, Aditya Siddhant, Aditya Barua, and Colin Raffel. 2021.
\newblock \href {https://doi.org/10.18653/v1/2021.naacl-main.41} {m{T}5: A massively multilingual pre-trained text-to-text transformer}.
\newblock In \emph{Proceedings of the 2021 Conference of the North American Chapter of the Association for Computational Linguistics: Human Language Technologies}, pages 483--498, Online. Association for Computational Linguistics.

\bibitem[{Yeh et~al.(2021)Yeh, Eskenazi, and Mehri}]{yeh-etal-2021-comprehensive}
Yi-Ting Yeh, Maxine Eskenazi, and Shikib Mehri. 2021.
\newblock \href {https://doi.org/10.18653/v1/2021.eancs-1.3} {A comprehensive assessment of dialog evaluation metrics}.
\newblock In \emph{The First Workshop on Evaluations and Assessments of Neural Conversation Systems}, pages 15--33, Online. Association for Computational Linguistics.

\bibitem[{Young(2007)}]{young2007cued}
Steve Young. 2007.
\newblock Cued standard dialogue acts.
\newblock \emph{Report, Cambridge University Engineering Department, 14th October}, 2007.

\bibitem[{Young(2010)}]{Young:2010}
Steve~J. Young. 2010.
\newblock \href {https://doi.org/10.1109/MSP.2010.935874} {Cognitive user interfaces}.
\newblock \emph{{IEEE} Signal Processing Magazine}, 27(3):128--140.

\bibitem[{Zhang et~al.(2023)Zhang, Peng, Li, Zhou, and Meng}]{zhang2023sgp}
Xiaoying Zhang, Baolin Peng, Kun Li, Jingyan Zhou, and Helen Meng. 2023.
\newblock \href {https://aclanthology.org/2023.findings-emnlp.891} {{SGP}-{TOD}: Building task bots effortlessly via schema-guided {LLM} prompting}.
\newblock In \emph{Findings of the Association for Computational Linguistics: EMNLP 2023}, pages 13348--13369, Singapore. Association for Computational Linguistics.

\bibitem[{Zhu et~al.(2022)Zhu, Geishauser, chin Lin, van Niekerk, Peng, Zhang, Heck, Lubis, Wan, Zhu, Gao, Gašić, and Huang}]{zhu2022convlab3}
Qi~Zhu, Christian Geishauser, Hsien chin Lin, Carel van Niekerk, Baolin Peng, Zheng Zhang, Michael Heck, Nurul Lubis, Dazhen Wan, Xiaochen Zhu, Jianfeng Gao, Milica Gašić, and Minlie Huang. 2022.
\newblock \href {http://arxiv.org/abs/2211.17148} {Convlab-3: A flexible dialogue system toolkit based on a unified data format}.
\newblock \emph{arXiv preprint arXiv:2211.17148}.

\bibitem[{Zhu et~al.(2020)Zhu, Zhang, Fang, Li, Takanobu, Li, Peng, Gao, Zhu, and Huang}]{zhu2020convlab2}
Qi~Zhu, Zheng Zhang, Yan Fang, Xiang Li, Ryuichi Takanobu, Jinchao Li, Baolin Peng, Jianfeng Gao, Xiaoyan Zhu, and Minlie Huang. 2020.
\newblock \href {https://doi.org/10.18653/v1/2020.acl-demos.19} {{C}onv{L}ab-2: An open-source toolkit for building, evaluating, and diagnosing dialogue systems}.
\newblock In \emph{Proceedings of the 58th Annual Meeting of the Association for Computational Linguistics: System Demonstrations}, pages 142--149, Online. Association for Computational Linguistics.

\end{thebibliography}
\bibliographystyle{acl_natbib}

\newpage

\appendix
\section{Experimental Details}
\label{sec:experiment_details}

In this section, we describe the experimental setups for the systems developed in this paper. For specific implementation details, including the prompts used in the ICL-based systems, we direct readers to the actual implementation and documentation of \toolkit.

\vspace{1.5mm}
\noindent \textbf{Table~\ref{tab:hyperparameters}} presents the selected hyper-parameters for the conducted experimental study. All the FT-based experiments were run on a single A100 80 GiB GPU and a 32-core vCPU. Notably, the ICL-based systems deployed in our experimental study excluded training examples from the prompts. This decision was based on empirical evidence indicating that these examples adversely affects system performance. When tested with 10 ICL examples, the ICL-GPT-3.5 systems recorded a JGA of 4.3 ($\downarrow$9.2), an Inform Rate of 31.0 ($\downarrow$2.0), and a Success Rate of 14.0 ($\downarrow$2.0). We did not conduct a hyperparameter search for the number of ICL examples due to the high costs associated with such an experiment.

\begin{table}[]
\centering
{\footnotesize
\begin{tabularx}{\linewidth}{l X}
\toprule
{\bf Hyper-parameter}                                   & {\bf Value}                         \\ \midrule
\multicolumn{2}{c}{\cellcolor[HTML]{EFEFEF}{\bf FT-mT5\textsubscript{small}}} \\ \midrule
batch size                              & 32                      \\
learning rate                           & 1e-3                     \\
weight decay                            & 0.01                      \\
evaluation per steps                    & 5000                      \\
max training steps                      & 50000                    \\
context window                          & 10                 \\
early stopping patience                 & 2                      \\
maximum generation length               & 512                      \\
\midrule
\multicolumn{2}{c}{\cellcolor[HTML]{EFEFEF} {\bf FT-mT5\textsubscript{large}}} \\ \midrule
batch size                              & 8                      \\
learning rate                           & 1e-3                     \\
weight decay                            & 0.01                      \\
evaluation per steps                    & 5000                      \\
max training steps                      & 50000                    \\
context window                          & 10                 \\
early stopping patience                 & 2                      \\
maximum generation length               & 512                      \\
\midrule
\multicolumn{2}{c}{\cellcolor[HTML]{EFEFEF} {\bf ICL-GPT-3.5}, {\bf ICL-LLaMA2}, and {\bf ICL-OpenChat-3.5}} \\ \midrule
context window                          & 10                 \\
number of ICL examples                  & 0 {\color{black}$(*)$}                \\
\bottomrule
\end{tabularx}
}%
\caption{The hyperparameters for E2E systems and their constituent models. Both the DST and RG models, which are based on the same PLM, utilised identical hyper-parameter setups. To select the optimal model checkpoint, we employ early stopping and select the one with the best validation performance, measured by JGA for DST and BLEU score for RG. Unless explicitly specified, all other hyper-parameters are set to their default values as defined in the HuggingFace Transformers. {\color{black}$(*)$} Notably, our observations suggest that the introduction of training examples adversely affects model performance.}
\label{tab:hyperparameters}
\end{table}

\vspace{1.5mm}
\noindent \textbf{Table~\ref{tab:input_plms}} lists all the language models we used in this work, along with their respective checkpoints in the Huggingface repository and the OpenAI API. Both the LLaMA2 and OpenChat-3.5 models employed in this study has 7 billion parameters.

\begin{table}[!t]
\centering

\begin{tabular}{@{}ll@{}}
\toprule
\textbf{Model}               & \textbf{Checkpoint}                     \\ \midrule
mT5\textsubscript{small}                 & google/mt5-small                    \\
mT5\textsubscript{large}                  & google/mt5-large                         \\
GPT-3.5                      & gpt-3.5-turbo-1106 \\
LLaMA2        & TheBloke/Llama-2-7B-GGUF  \\
OpenChat-3.5 & openchat/openchat\_3.5  
\\\bottomrule
\end{tabular}
\caption{The employed langauge models in our experimental study and their Huggingface or OpenAI Checkpoints. We use 7B variants of LLaMA2 and OpenChat-3.5.}
\label{tab:input_plms}
\end{table}

\vspace{1.5mm}
\noindent \textbf{Table~\ref{tab:time}} shows the time consumption of the models for the E2E task in the experimental study.

\begin{table}[!t]
\centering
{\footnotesize
\begin{tabularx}{\linewidth}{l X}
\toprule
\textbf{Setup}                                    & \textbf{Time Consumption }            \\ \midrule
\multicolumn{2}{c}{\cellcolor[HTML]{EFEFEF} {\bf FT-mT5\textsubscript{small}} } \\ \midrule
DST training per 500 steps                              & 3:02                        \\
RG training per 500 steps                              & 2:27                        \\
Inference on \textit{full test}                        & 9:30                         \\ \midrule
\multicolumn{2}{c}{\cellcolor[HTML]{EFEFEF} {\bf FT-mT5\textsubscript{large}} } \\ \midrule
DST training per 500 steps                              & 5:09                        \\
RG training per 500 steps                              & 5:02                        \\
Inference on \textit{full test}                        & 1:16:00            \\  \midrule
\multicolumn{2}{c}{\cellcolor[HTML]{EFEFEF} {\bf ICL-GPT-3.5}} \\ \midrule
Inference on \textit{10 dialogues}                        & 9:05                         \\ \midrule
\multicolumn{2}{c}{\cellcolor[HTML]{EFEFEF}{\bf ICL-LLaMA2}} \\ \midrule
Inference on \textit{10 dialogues}                        & 36:00                         \\ \midrule
\multicolumn{2}{c}{\cellcolor[HTML]{EFEFEF}{\bf ICL-OpenChat-3.5}} \\ \midrule
Inference on \textit{10 dialogues}                        & 5:35                         \\ 
\bottomrule
\end{tabularx}
}%
\caption{The average time consumption for the E2E task. For all FT-based systems, the computation was performed on a machine equipped with a single A100 80 GiB GPU and a 32-core vCPU. In the case of the ICL-GPT-3.5 systems, calculations were conducted using the OpenAI API. Meanwhile, the ICL-LLaMA2 systems were executed on an Intel 13900k CPU and  ICL-OpenChat-3.5 systems were executed on a machine with an Intel 13900k CPU and a single NVIDIA RTX 4090 GPU.}
\label{tab:time}
\end{table}

\vspace{1.5mm}
\noindent \textbf{Table~\ref{tab:auto_eval_appendix}} shows the fully supervised performance of LLaMA2 and OpenChat-3.5 models across DST models, RG models, and E2E systems on the \dataset dataset. It is noteworthy that the OpenChat-3.5 based systems exhibited a failure in generating coherent dialogue responses. These systems consistently repeated the prompts and, in each utterance, indiscriminately included all the placeholders. This simplistic method led to inflated Inform Rate and Success Rate scores, highlighting the potential vulnerability of these metrics to adversarial strategies, a concern also highlighted by~\citet{wu2023using}.

\begin{table*}[]
\centering
\resizebox{0.7\textwidth}{!}{%
\begin{tabular}{@{}lccccclccclccc@{}}
\toprule
                           &                      & \multicolumn{4}{c}{\textbf{Dialogue State Tracking}}                                & \multicolumn{1}{c}{} & \multicolumn{3}{c}{\textbf{Response Generation}} &  & \multicolumn{3}{c}{\textbf{End-to-end Modelling}} \\ \cmidrule(lr){3-6} \cmidrule(lr){8-10} \cmidrule(l){12-14} 
\multirow{-2}{*}{Language} &                      & JGA                  & Slot F1              & Slot Precision          & Slot Recall &                      & BLEU          & ROUGE          & METEOR          &  & Inform Rate       & Success Rate      & BLEU      \\ \midrule
\multicolumn{14}{c}{\cellcolor[HTML]{EFEFEF}{\color[HTML]{000000} ICL-LLaMA2 {\color{black}$(*)$}}}                                                                                                                                                                                            \\ \midrule
ENG                     &    &    1.6               &     0.0                 &           0.0           &        0.0                   &                      &    0.1           &  0.1              &          0.1       &  &     22.0          &        8.0        &  0.2        \\
ARA                        &                      &                1.6               &     0.0                 &           0.0           &        0.0            &                      &      0.1      &      0.0      &       0.0      &  &        23.0        &     9.0          &      0.1     \\
FRA                        &                      &                    1.6                &     0.0                 &           0.0           &        0.0             &                      &     0.0       &     0.0      &       0.0        &  &    19.0        &           9.0   &   0.0   \\
TUR                        &                      &       1.6     &   0.0              &       0.0         &      0.0                &                      &    0.0       &   0.0        &       0.0  &  &       18.0      &     7.0         &  0.0     \\
AVG.                       &                      &        1.6     &   0.0              &       0.0         &      0.0      &                      &     0.1        &  0.0       &        0.0       &  &       20.5      &     8.3         &  0.0   \\ 
\midrule
\multicolumn{14}{c}{\cellcolor[HTML]{EFEFEF}{\color[HTML]{000000} ICL-OpenChat-3.5 {\color{black}$(*)$} {\color{black}$(**)$}}}                                                                                                                                                                                            \\ \midrule
ENG                        &   &  1.6 & 0.0 & 0.0 &       0.0         &                      &      0.0         &       0.0         &  0.0              &  &       67.0         &       61.0         &    0.0       \\
ARA                        &  &  1.6 & 0.0 & 0.0 &       0.0       &                      &     0.0       &        0.0       &      0.0           &  &        67.0        &        60.0           &     0.0      \\
FRA                        &   & 1.6 & 0.0 & 0.0 &       0.0         &                      &     0.0          &      0.0         &    0.0             &  &        67.0         &       60.0            &   0.0        \\
TUR                        &   &   1.6  &  0.0  & 0.0 &   0.0          &                      &     0.0          &   0.0           &   0.0              &  &             67.0      &     60.0             &     0.0      \\
AVG.                       &   & 1.6 & 0.0 &   0.0             &        0.0             &   &      0.0        &        0.0        &      0.0           &  &     67.0              &    60.3         &   0.0        \\ 

\bottomrule
\end{tabular}%
}
\caption{Evaluation of fully supervised performance across DST models, RG models, and E2E systems on the \dataset dataset. This table reports the performance metrics for each language, evaluated across different models. It should be noted that for these metrics, the ground truth score is set at 100, with the exceptions of the Inform Rate and Success Rate, which are measured as $89.3\pm0.2$ and $68.6\pm0.2$ across the four languages, respectively. $(*)$ For practical considerations, the evaluation ICL-based models and systems is limited to a randomly selected sample of 100 dialogues from the full test set, due to the significant time and resource requirements of a full-scale evaluation. 
$(**)$ Additionally, it is noteworthy that the OpenChat-3.5 based systems exhibited a failure in generating coherent dialogue responses. These systems repetitively echoed the prompts and for each utterance, generated all the placeholders. This simplistic approach resulted in artificially high Inform Rate and Success Rate scores, revealing the vulnerability of these metrics to adversarial strategies.}
\label{tab:auto_eval_appendix}
\end{table*}

\noindent \textbf{Table~\ref{tab:auto_eval_100}} presents an analysis of the performance under full supervision for both mT5\textsubscript{small} and mT5\textsubscript{large} models. This evaluation is conducted on a specifically selected subset of 100 dialogues from the entire test set, consistent with the evaluation setup applied to all other ICL-based models and systems in this study. This approach ensures a rigorous and direct comparability across the discussed ICL-based models and systems.

\begin{table*}[]
\centering
\resizebox{0.7\textwidth}{!}{%
\begin{tabular}{@{}lccccclccclccc@{}}
\toprule
                           &                      & \multicolumn{4}{c}{\textbf{Dialogue State Tracking}}                                & \multicolumn{1}{c}{} & \multicolumn{3}{c}{\textbf{Response Generation}} &  & \multicolumn{3}{c}{\textbf{End-to-end Modelling}} \\ \cmidrule(lr){3-6} \cmidrule(lr){8-10} \cmidrule(l){12-14} 
\multirow{-2}{*}{Language} &                      & JGA                  & Slot F1              & Slot Precision          & Slot Recall &                      & BLEU          & ROUGE          & METEOR          &  & Inform Rate       & Success Rate      & BLEU      \\ \midrule
\multicolumn{14}{c}{\cellcolor[HTML]{EFEFEF}FT-mT5\textsubscript{small}{\color{black}$(*)$}}                                                                                                                                                                                                 \\ \midrule
ENG                     &    &   54.4                &     83.0                &           84.6           &        81.5                   &                      &    18.7           &  26.5             &          29.7       &  &     66.0          &        47.0        &  18.5        \\
ARA                        &                      &                41.9                &     77.5                 &           79.9           &        75.3            &                      &      16.1      &      30.0      &       27.4      &  &        67.0         &     43.0          &      15.8     \\
FRA                        &                      &                    42.9                &     80.3                 &           81.3           &        79.3             &                      &    12.9     &    25.0      &      25.6        &  &    66.0        &            34.0   &    13.4   \\
TUR                        &                      &       48.0                &     81.5                 &           81.1           &        81.9               &                      &    22.8       &   34.6        &       34.1  &  &       75.0        &       48.0        &    23.1    \\
AVG.                       &                      &        46.8      &   80.6               &       81.7        &      79.5      &                      &     17.6        &  29.0       &        29.2       &  &       68.5      &     43.0         &  17.7   \\ \midrule
\multicolumn{14}{c}{\cellcolor[HTML]{EFEFEF}FT-mT5\textsubscript{large}{\color{black}$(*)$}}                                                                                                                                                                                                 \\ \midrule
ENG                        &                      &    19.1       &      53.3           &    55.0           &    51.6       &                      &      18.0     &      25.7          &       29.0          &  &       69.0       &           49.0        &      17.8     \\
ARA                        &  &  41.9 &  79.1 & 80.2 &  78.1              &                      &        9.2       &   19.4             &     17.0            &  &  62.0                &    30.0            &  9.0          \\
FRA                        &                      &        44.6              &           80.3           &             82.1         &       78.5         &       &      13.1              &       24.4        &     24.6                         &  &               77.0         &   50.0          &     13.4      \\
TUR                        &             &      45.2                    &      80.8          &     80.8                 &     80.8      &                      &       11.2      &     20.5           &        19.7         &  &         68.0     &    28.0             &      11.4 \\
AVG.                       &                      &   30.2             &      73.4              &      74.5               &   72.3           &                      &      12.9         &       22.5         &      22.6           &  &   69.0         &   39.3          &     12.9    \\ 
\bottomrule
\end{tabular}%
}
\caption{Evaluation of fully supervised performance across DST models, RG models, and E2E systems on the \dataset dataset. This table reports the performance metrics for each language, using both mT5\textsubscript{small} and mT5\textsubscript{large} models. `AVG.' represents the mean average of the evaluation scores aggregated across all four languages. We note that for these metrics the ground truth score is set at 100, with the exception of the Inform Rate and Success Rate, which are measured as $89.3\pm0.2$ and $68.6\pm0.2$ across the four languages, respectively. {\color{black}$(*)$} In this table, the evaluation is limited to a randomly selected sample of 100 dialogues from the full test set, ensuring direct comparability with other ICL-based models and systems discussed herein.}
\label{tab:auto_eval_100}
\end{table*}

\section{Screenshots}
\label{sec:screenshots}

In this section, we provide screenshots of our web interface alongside implementations of our toolkit. These figures are intended to illustrate the essential features and functionalities of our toolkit.

\begin{figure}[!t]
    \centering
    \includegraphics[width=\linewidth]{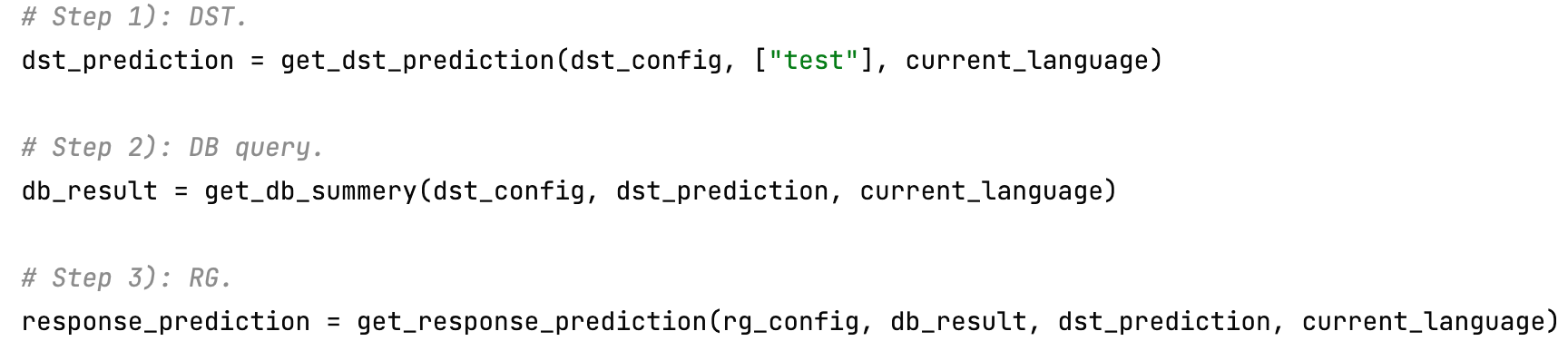}
    \caption{A screenshot of the inference code for our FT-based E2E systems is provided above. The code is intentionally designed to be both simple and fully functional, aiming to assist users in gaining a conceptual understanding of the \tod task and the implementation of our system.}
    \label{fig:e2e_code}
\end{figure}

\vspace{1.5mm}
\noindent \textbf{Figure~\ref{fig:e2e_code}} presents a screenshot capturing the login page of our human evaluation web interface. This interface serves as the entry point for evaluators to access the system.

\begin{figure}[!t]
    \centering
    \includegraphics[width=\linewidth]{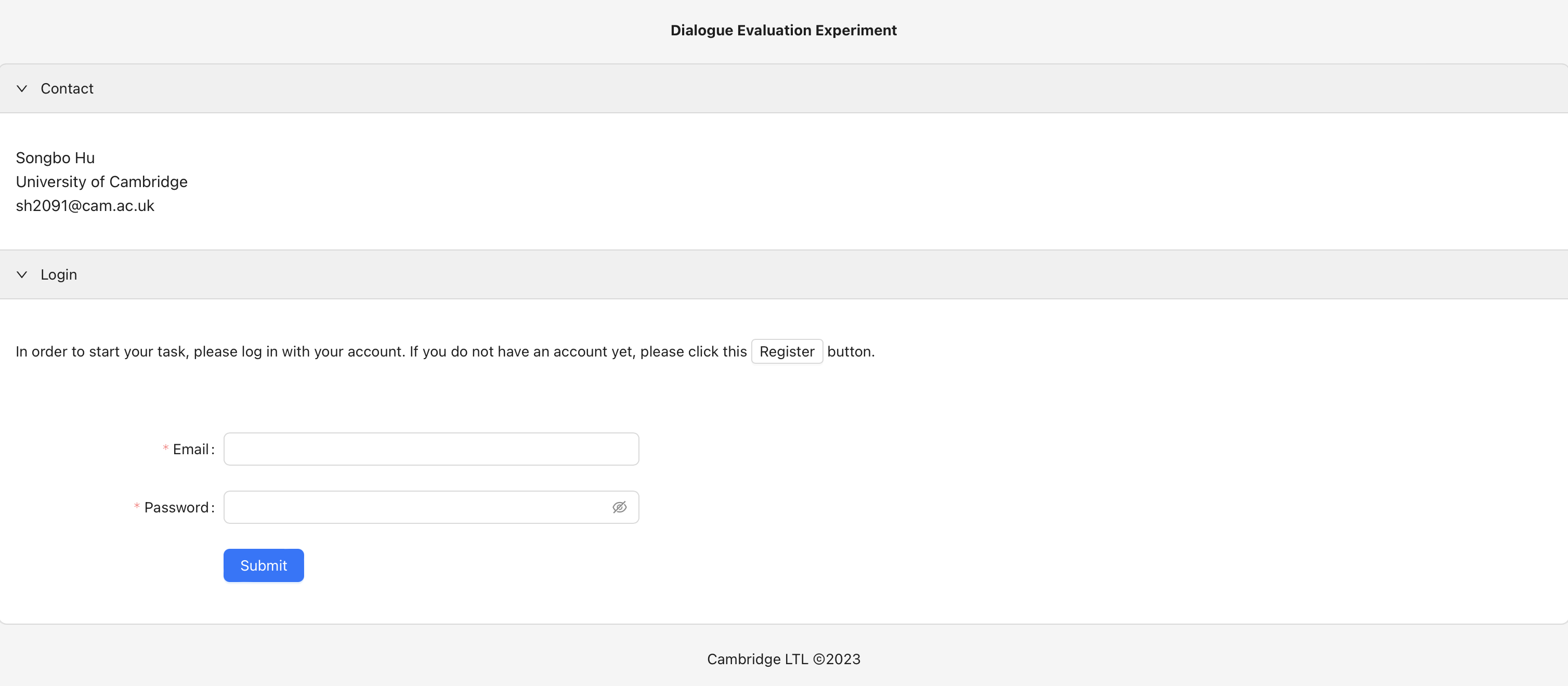}
    \caption{A screenshot capturing the login page of the human evaluation web interface.}
    \label{fig:login}
\end{figure}

\begin{figure}[!t]
    \centering
    \includegraphics[width=\linewidth]{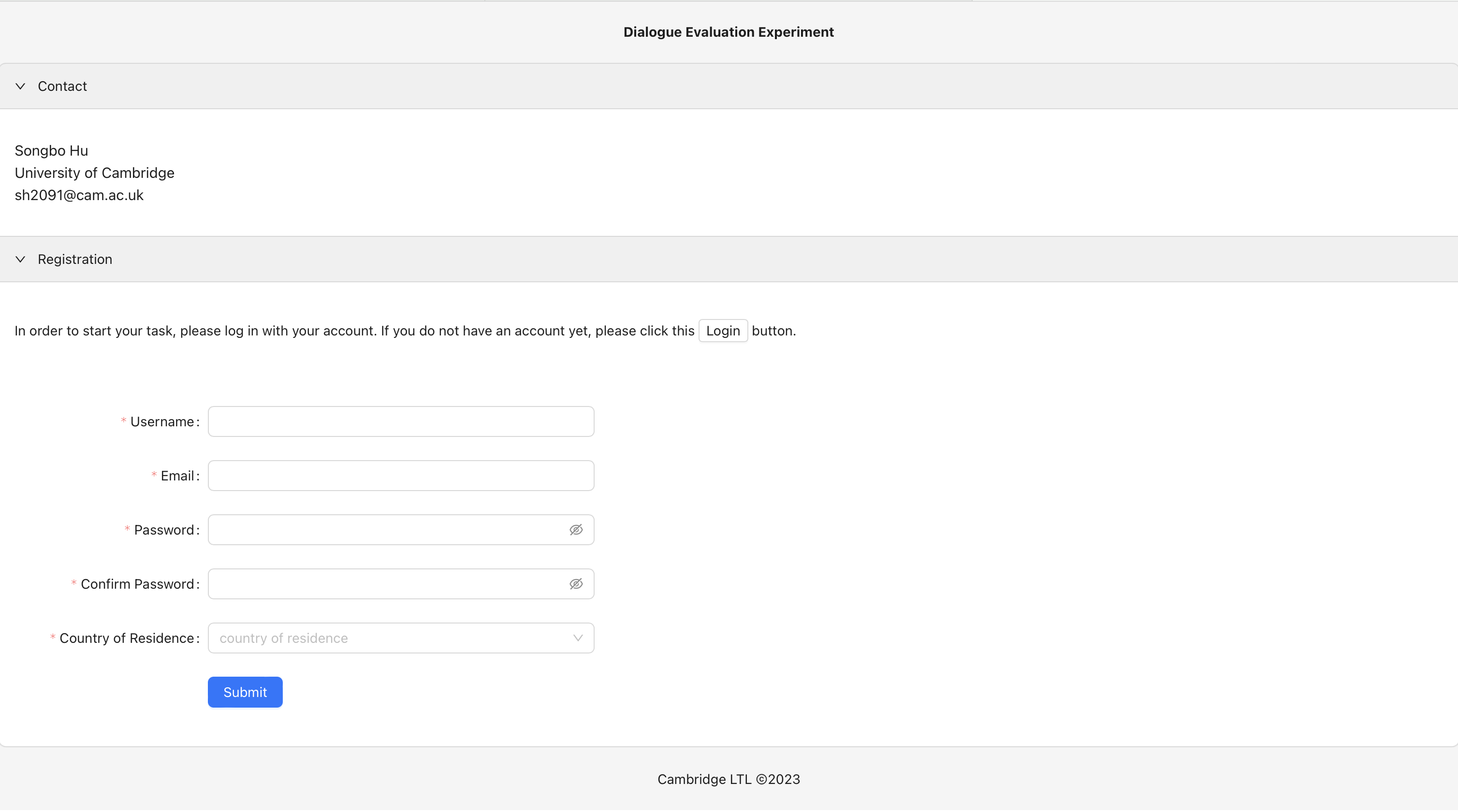}
    \caption{A screenshot of the registration page.}
    \label{fig:register}
\end{figure}

\vspace{1.5mm}
\noindent \textbf{Figure~\ref{fig:register}} presents a screenshot capturing the registration page of our human evaluation web interface.

\begin{figure}[!t]
    \centering
    \includegraphics[width=\linewidth]{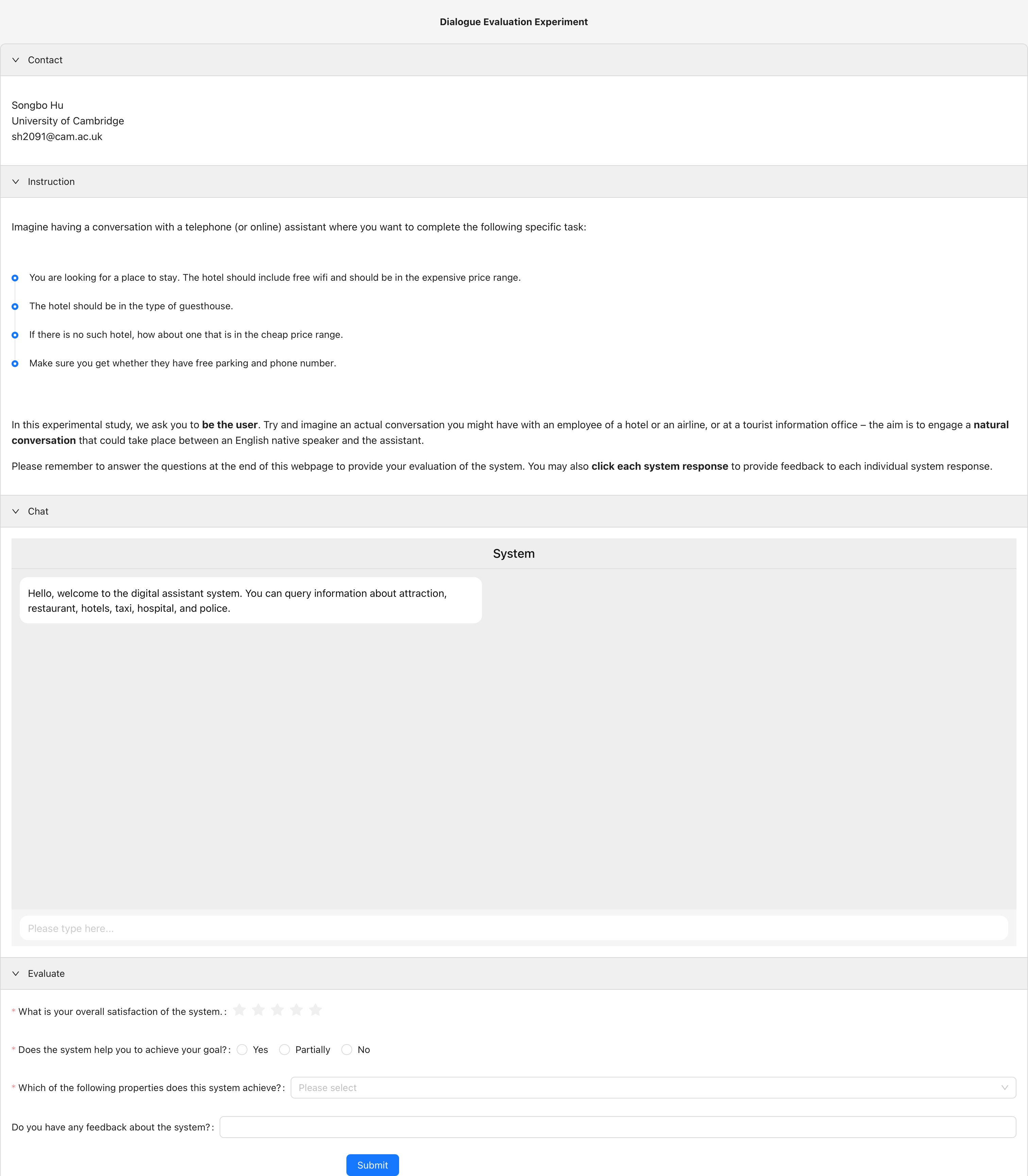}
    \caption{A screenshot of the task assignment page.}
    \label{fig:task}
\end{figure}

\vspace{1.5mm}
\noindent \textbf{Figure~\ref{fig:task}} displays a screenshot of the assignment page within the human evaluation web interface. This page is specifically designed for users to carry out the task of evaluating the dialogue system.

\vspace{1.5mm}
\noindent \textbf{Figure~\ref{fig:e2e_code}} presents a screenshot of the inference code for our FT-based E2E systems. The code is modularised and intentionally designed to be both simple and fully functional. Our goal is to facilitate users in acquiring a clear understanding of the \tod task, as well as to provide insights into the implementation of our system.

\begin{figure}[!t]
    \centering
    \includegraphics[width=\linewidth]{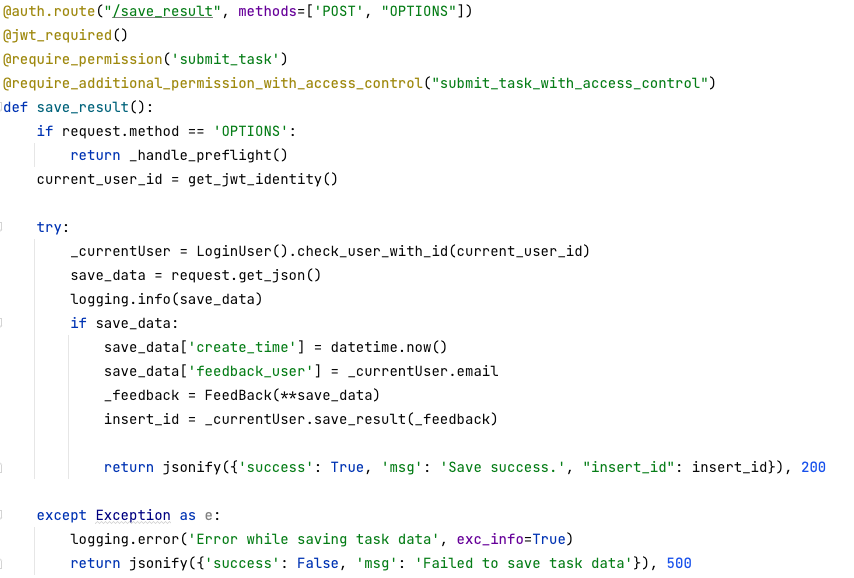}
    \caption{A screenshot of the backend web server code for our RESTful API, designed for storing system evaluation results. It features a JWT-based authentication mechanism to ensure secure access. Furthermore, the system is configured to allow only users with specific permissions to save evaluation results to the database, thereby enhancing data integrity and security.
}
    \label{fig:api}
\end{figure}

\vspace{1.5mm}
\noindent \textbf{Figure~\ref{fig:api}} presents a screenshot of the backend web server code for our RESTful API for the storage of human evaluation results. This setup incorporates a JWT-based authentication system to secure access. Additionally, it is structured to permit only authorised users with specific permissions to record evaluation results in the database.

\end{document}